%% file: 0_main.tex
\pdfoutput=1
\documentclass[11pt]{article}

\usepackage[preprint]{acl}

\usepackage{times}
\usepackage{latexsym}
\usepackage{caption}
\usepackage{subcaption}

\usepackage[T1]{fontenc}
\usepackage[utf8]{inputenc}

\usepackage{microtype}

\usepackage{inconsolata}

\usepackage{graphicx}
\usepackage{multirow, booktabs}
\usepackage{siunitx}
\usepackage{amssymb}
\usepackage{arydshln}
\usepackage{xcolor}
\title{Towards Better Multi-task Learning: A Framework for Optimizing Dataset Combinations in Large Language Models}

\author{
 \textbf{Zaifu Zhan\textsuperscript{1}},
 \textbf{Rui Zhang\textsuperscript{2}}
\\
 \textsuperscript{1}Department of Electrical and Computer Engineering, University of Minnesota, Minneapolis, MN, USA
 \\
 \textsuperscript{2}Division of Computational Health Sciences, Department of Surgery, University of Minnesota, \\Minneapolis, MN, USA
\\
 \small{
   \textbf{Correspondence:} Rui Zhang, PhD \href{(zhan1386@umn.edu)}{(zhan1386@umn.edu)}
 }
}

\begin{document}
\maketitle
\begin{abstract}
To efficiently select optimal dataset combinations for enhancing multi-task learning (MTL) performance in large language models, we proposed a novel framework that leverages a neural network to predict the best dataset combinations. The framework iteratively refines the selection, greatly improving efficiency, while being model-, dataset-, and domain-independent. Through experiments on 12 biomedical datasets across four tasks—named entity recognition, relation extraction, event extraction, and text classification—we demonstrate that our approach effectively identifies better combinations, even for tasks that may seem unpromising from a human perspective. This verifies that our framework provides a promising solution for maximizing MTL potential.
\end{abstract}

\input{2_introduction}
\input{3_related_work}
\input{4_preliminaries}
\input{5_method}

\input{6_experiment}

\input{7_conclution}

% \section*{Acknowledgments}

\bibliography{custom}

\input{8_appendix}

\end{document}

%% file: 2_introduction.tex
\section{Introduction}

Natural Language Processing (NLP) has made significant strides in recent years~\cite{liu2023pre}, evolving from \textit{fully supervised learning} ~\cite{kotsiantis2007supervised}, to \textit{feature engineering} ~\cite{patil2023survey}, \textit{architecture innovations} like Transformer ~\cite{vaswani2017attention}, and the dominance of \textit{pre-trained large models} such as BERT and GPT ~\cite{devlin2018bert, radford2018improving, radford2019language}. More recently, the \textit{instruction-tuning} ~\cite{zhang2023instruction} and \textit{prompting engineering} ~\cite{liu2023pre} have emerged, allowing Large Language Models (LLMs) to handle tasks effectively through prompting~\cite{wei2022chain}.

\begin{figure}[t]
  \centering
  \begin{subfigure}[t]{0.48\linewidth}
    \centering
    \includegraphics[width=\linewidth]{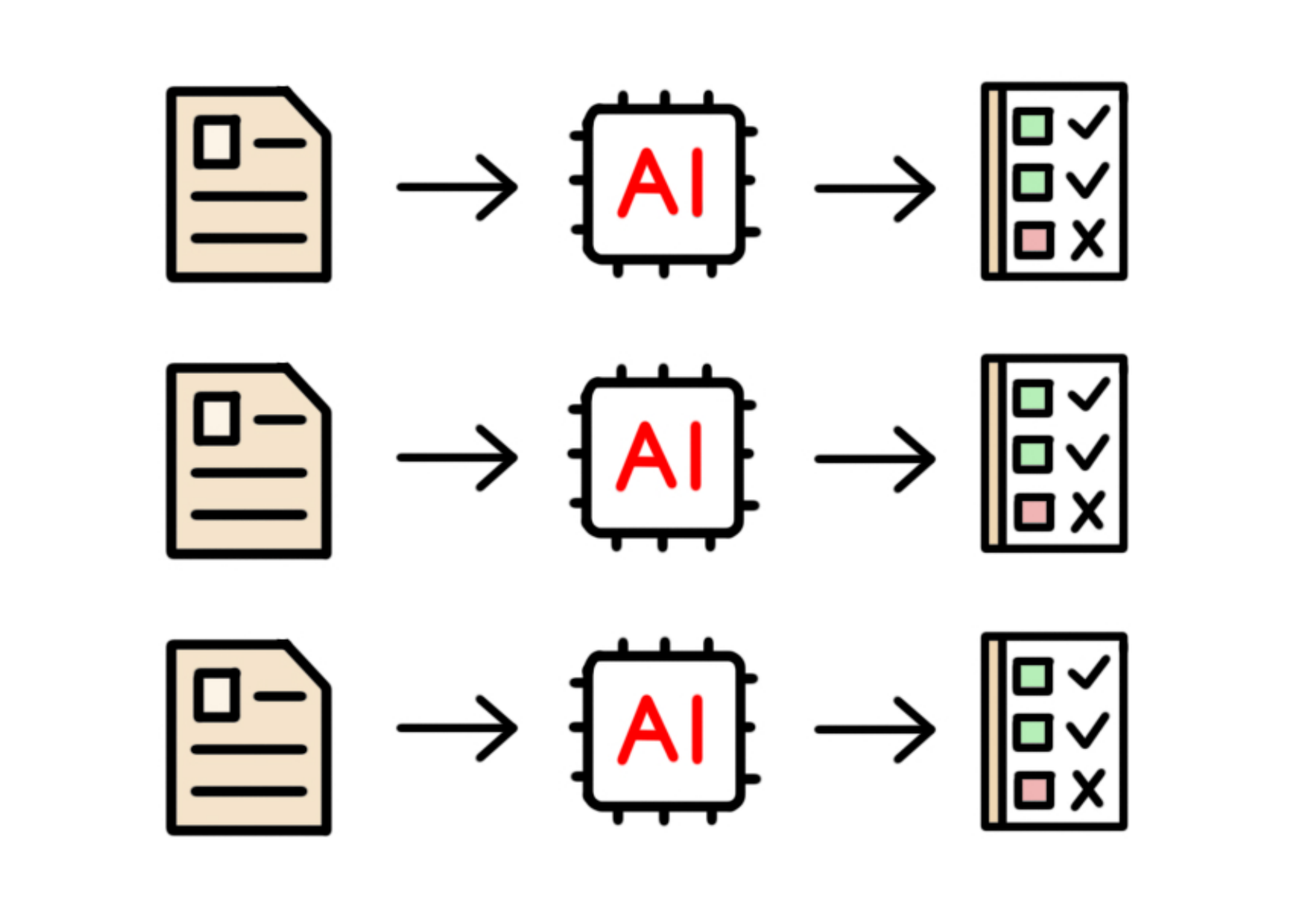}
    \caption{Single-task learning}
    \label{fig:stl}
  \end{subfigure}
  \hfill
  \begin{subfigure}[t]{0.45\linewidth}
    \centering
    \includegraphics[width=\linewidth]{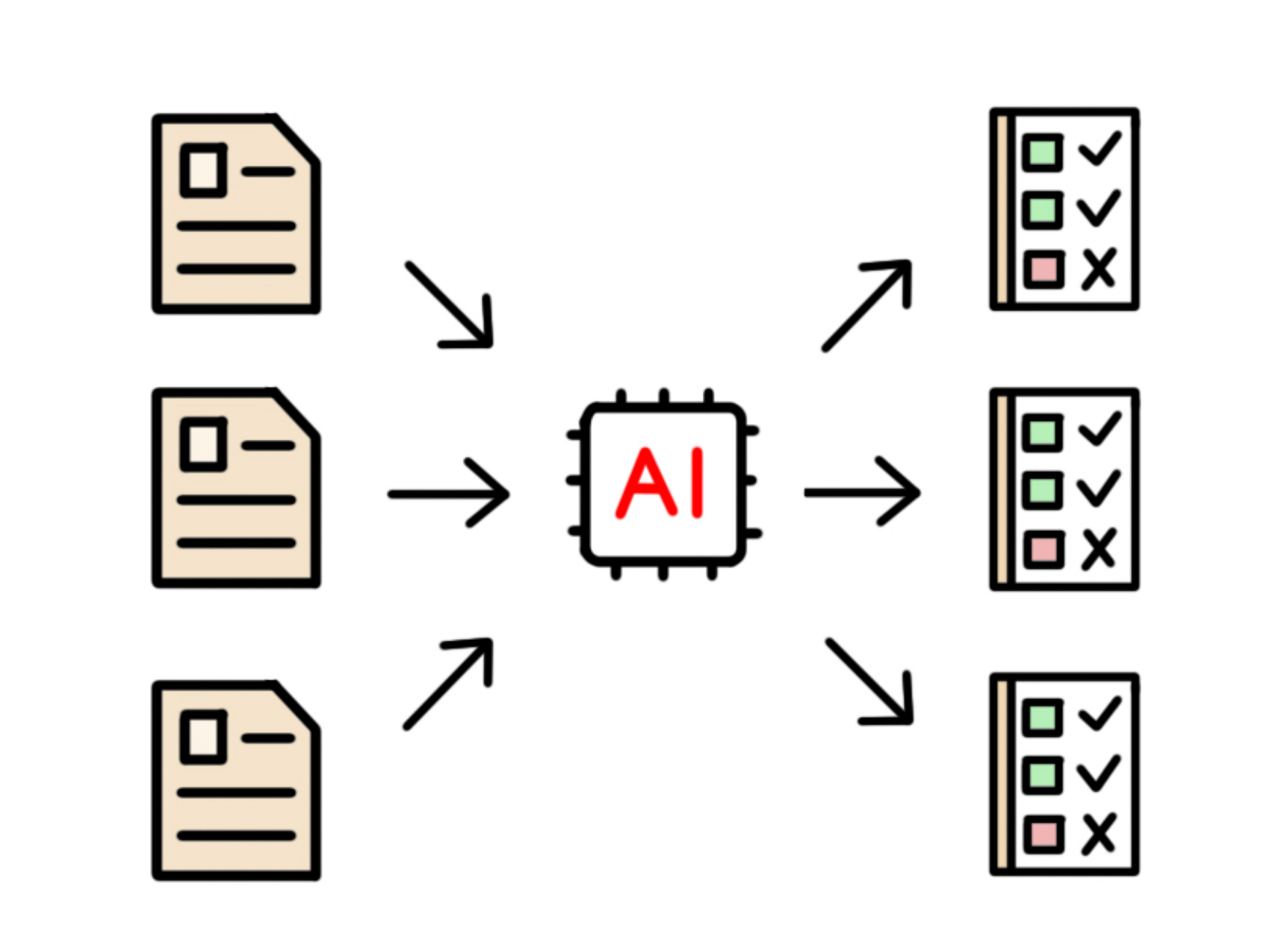}
    \caption{Multi-task learning}
    \label{fig:mtl}
  \end{subfigure}
  \caption{Comparison of (a) single-task learning and (b) multi-task learning in large language models.}
  \label{fig:comparison}
\end{figure}

With the rapid advancements in NLP, Multi-Task Learning (MTL) has emerged as a powerful technique to boost model performance by jointly training on multiple related tasks ~\cite{zhang2018overview, zhang2021survey}, as illustrated in Fig. \ref{fig:comparison}. By sharing knowledge across tasks, MTL enhances model generalization~\cite{wang2021bridging} and efficiently captures the complementary relationships between tasks~\cite{ma2018modeling}. For instance, the Named Entity Recognition (NER) task and the Relation Extraction (RE) task are closely linked—accurate entity recognition can provide a critical context for extracting relationships, while relation extraction, in turn, can refine entity boundaries and classifications.

As models become larger, the amount of data used for training has also significantly increased~\cite{chen2024multi}, giving rise to models capable of understanding and solving problems across various domains. 
For instance, ChatGPT can generate realistic and creative outputs across various domains~\cite{yenduri2024gpt}. 
Previously, achieving MTL required modifying the model architecture and adjusting the output layers for different tasks~\cite{misra2016cross}. 
Now, by simply modifying the prompt, MTL has found greater utility and flexibility without re-designing the architecture~\cite{liu2023pre}.
This indicates that large models have truly become tools accessible to everyone. As long as users know how to utilize frameworks like Hugging Face~\cite{wolf-etal-2020-transformers}, they can fine-tune models with their own prompts without needing any knowledge of Attention mechanisms or model architecture.

To take advantage of the natural compatibility of MTL and LLMs, many recent large models have improved performance by incorporating multi-task training. Models like DeepStruct ~\cite{wang2022deepstruct}, InstructUIE ~\cite{wang2023instructuie}, and Code4Struct ~\cite{wang2022code4struct} have demonstrated the power of multi-task learning by training on diverse datasets across tasks like NER, RE, Event Extraction (EE), and Slot Filling (SF), etc. In addition, GoLLIE ~\cite{sainz2023gollie} used datasets spanning both biomedical and news domains, achieving state-of-the-art results in these fields. ADELIE ~\cite{qi2024adelie} further emphasized that aligning LLMs with multiple Information Extraction (IE) tasks significantly improves task performance in the biomedical domain.

\begin{figure*}[t]
    \centering
    \includegraphics[width=0.9\linewidth]{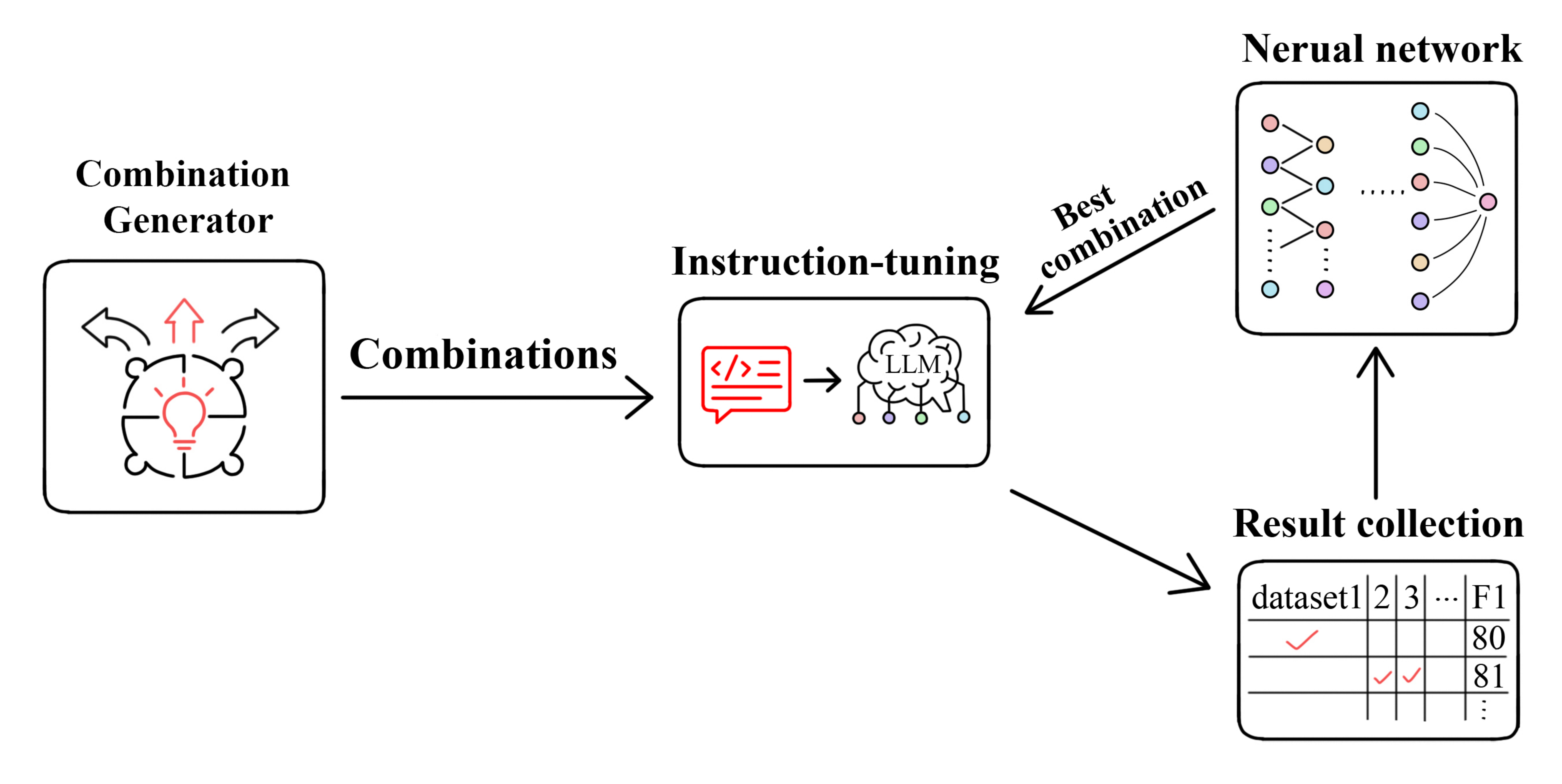}
    \caption{\textbf{Overview of the proposed framework.} The combination generator produces dataset combinations to be instruction-tuned on an LLM. In each iteration, the neural network identifies and refines the best combination until no further improvements can be made.}
    \label{fig:2_framework}
\end{figure*}

Despite the encouraging results achieved by many multi-task LLMs, we still lack a clear understanding of how to effectively select datasets for training. Most successful outcomes rely heavily on trial and error, subjective judgment, or experience.
This reveals a significant question: how to effectively select dataset combinations to enhance performance?
If we have one dataset and four related auxiliary datasets, there are $2^4=16$ possible combinations, which is feasible to enumerate. However, when the number of auxiliary datasets increases to 10 or more, it becomes practically impossible to train and evaluate all possible combinations to find the optimal one.

To take a step toward filling the gap, this paper proposes a framework to identify a good combination of datasets for improving model performance.
The framework is dataset-independent and model-independent, so the framework could be applied to any domain, datasets, and to any LLMs.
The full description of the proposed framework is in Section \ref{sec.fw}.
% Section.\ref{} presents the results.
The main contributions of this paper are as follows: 
\begin{itemize} 
\item We propose a new MTL framework designed specifically to effectively optimize the selection of datasets to release the potential of MTL-LLMs. 
\item Based on this framework, we conduct a comprehensive evaluation of 12 biomedical datasets across four tasks: NER, EE, RE, and TC. The results show that the performance of LLMs on datasets could be increased by finding better combinations using our framework.
\end{itemize}

% Biomedical NLP, in particular, has benefited from the advancements of LLMs. This field focuses on extracting and analyzing information from vast amounts of biomedical text ~\cite{yang2023large}, which is essential for applications like drug discovery~\cite{pal2023chatgpt}, clinical decision-making~\cite{zhou2024large}, and personalized medicine ~\cite{monajatipoor2024llms}. However, biomedical texts present unique challenges, including complex terminology, abbreviations, and ambiguities. General-purpose LLMs, despite their strengths, often struggle with these complexities without additional fine-tuning or continual learning ~\cite{ke2021achieving, hu2021lora, chen2024evaluating}.
% Therefore, 

%% file: 3_related_work.tex
\section{Related Work}
There have been several attempts to explore how to select dataset combinations to improve MTL~\cite{zhang2018overview,zhang2021survey,thung2018brief,sener2018multi,crawshaw2020multi}.
For instance, ~\cite{bingel2017identifying} systematically investigated and found the MTL gains are related to the characteristics and features of datasets.
However, what kind of characteristics and features we should consider for different domains are unknown, so the analysis is mostly based on experience.
Also, ~\cite{standley2020tasks} tried to figure out which tasks should be learned together by considering the space of all possible task subsets, training networks for each subset, and then using each network’s performance to choose the best combination. 
It is straightforward, but training all possible combinations is impossible when we consider many tasks.
~\cite{pruksachatkun2020intermediate} perform a large-scale study
on the pre-trained RoBERTa model with 110 intermediate–target task combinations, and then evaluate all trained models with 25 probing tasks.
However, they failed to observe more granular correlations between probing and target task performance.
~\cite{guo2019autosem} proposed AUTOSEM which uses a multi-armed bandit controller to find appropriate auxiliary tasks but each arm only considers the relation between 2 tasks and doesn't consider the interaction of two or more auxiliary tasks.
~\cite{fifty2021efficiently} proposed to measure inter-task affinity by training all tasks together in a single multi-task network and quantifying the effect to which one task gradient update would affect another task loss.
It effectively computes task groupings from only a single training run but we know the gradient update depends on the loss function and the initial point.
For non-convex loss functions, there may be many local optima, so the grouping results from this method are various.

By contrast, our proposed framework avoids unreliable human experience and low effective brute force, using a simple neural network to find good dataset combinations based on some combination-score pair data.

%% file: 4_preliminaries.tex
\section{Framework}\label{sec.fw}
The most straightforward approach to evaluate the effect of all possible dataset combinations on a given LLM would be to directly do experiments for each combination via fine-tuning and inference of the corresponding LLM. 
However, this naive method is computationally expensive and requires substantial resources. 
By contrast, multi-layer neural networks are much faster to compute. If a fast neural network could effectively filter out combinations that are unlikely to yield good results, or directly predict the best combination, it would save significant time and computational power. Driven by this idea, we propose a new framework to optimize dataset selection for MTL.

As shown in Fig. \ref{fig:2_framework}, our proposed framework consists of four parts. First, we generate sufficient combinations of datasets to at least cover all datasets. Next, fine-tuning LLMs on these dataset combinations. After inference for each combination, we record all combinations and their corresponding performance scores in a table. 
Using this table data, we then train a neural network to predict the best dataset combination by enumerating all possible combinations through the neural network, which could predict the best combination in a super-efficient way. 
The framework iteratively trains the large model with the predicted combinations and tests the results. 
The process continues until the neural network predicts an optimal combination that has already been tested, at which point the loop terminates.

The proposed framework offers several advantages. First, it saves time by using a neural network to infer relationships between datasets, allowing us to focus on testing the most promising combinations based on existing data. Second, it is highly flexible, applicable across different models and dataset pools, and adaptable to various neural network architectures for predicting the optimal combination. Lastly, it is robust. Even if the initial selections are limited or random, and the neural network struggles to predict good combinations due to insufficient data, the iterative process ensures that as more data is gathered, the system will eventually find a good (or even optimal) combination.

The inspiration for this framework comes from the concept of feedback in control systems~\cite{doyle2013feedback}. In a data-driven system, finding the optimal controller requires first accumulating a certain amount of data and then trying to control the systems using the controller trained by the accumulated data~\cite{kiumarsi2014reinforcement}. With each attempt, more feedback is gathered, allowing the system to refine the controller. The better the controller becomes, the closer it gets to finding the optimal solution. Similarly, in our framework, we begin by generating various dataset combinations and get the corresponding performance score for each combination by inference. The neural network, much like a feedback-driven controller, gives lower scores to poor combinations and higher scores to promising ones. As the process continues, the neural network encourages the exploration of good combinations and discourages poor ones. With more iterations, the system converges towards identifying the optimal dataset combination.

%% file: 5_method.tex
\begin{figure*}
    \centering
    \includegraphics[width=0.9\linewidth]{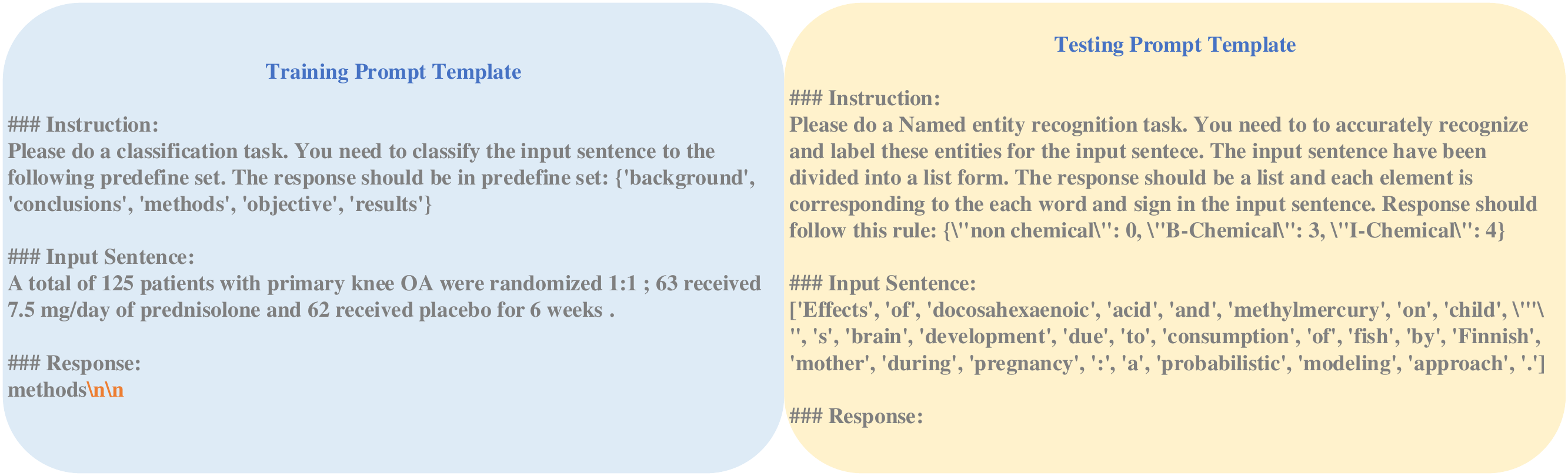}
    \caption{Training and testing prompt template. }
    \label{fig:3_prompt}
\end{figure*}

\section{Experiment setup}
\subsection{Dataset}
In this paper, we focus on four critical NLP tasks: EE, RE, NER, and TC. For each of these tasks, we selected three datasets to evaluate the potential of our framework comprehensively.

For \textbf{Event Extraction} task, we utilize the following datasets:
\begin{itemize}
    \item \textbf{PHEE} ~\cite{sun2022phee}: A comprehensive dataset providing high-accuracy annotations for 2 events: 'Adverse\_event', 'Potential\_therapeutic\_event'.
    \item \textbf{GENIA2011} ~\cite{kim2011overview}: A benchmark EE dataset from the biomedical domain. The event types include
    'Regulation', 'Localization', 'Transcription', 'Binding', 'Gene\_expression', 'Positive\_regulation', 'Protein\_catabolism', 'Negative\_regulation', 'Phosphorylation' and 'NA', totally 10 event types.
    \item \textbf{GENIA2013} ~\cite{kim2013genia}: An updated version of the genia2011 dataset, with additional and refined annotations reflecting the latest biomedical research. The event type include 'Regulation', 'Negative\_regulation', 'Protein\_modification', 'Positive\_regulation', 'Localization', 'Phosphorylation', 'Ubiquitination', 'Protein\_catabolism', 'Gene\_expression', 'Binding', 'Transcription' and 'NA, totally 12 event types.
\end{itemize}

For \textbf{Relation Extraction} task, we employ:
\begin{itemize}
    \item \textbf{DDI} ~\cite{segura2013semeval}: The Drug-Drug Interaction dataset, crucial for identifying interactions between different medications. The relation types include  'advise', 'effect', 'int', 'mechanism', and 'NA', totally 5 relation types.
    \item \textbf{GIT} ~\cite{li2023petailor}: A dataset focused on general interactions between entities, aiding in the development of versatile relation extraction models.
    There are 22 relation types: 'PREVENTS', 'TREATS', 'DOES\_NOT\_TREAT', 'ASSOCIATED\_WITH', 'CAUSES', 'DIAGNOSES', 'MANIFESTATION\_OF', 'USES', 'STIMULATES', 'INHIBITS', 'DISRUPTS', 'INTERACTS\_WITH', 'PRODUCES', 'ADMINISTERED\_TO', 'COEXISTS\_WITH', 'AFFECTS', 'PROCESS\_OF', 'COMPLICATES', 'AUGMENTS', 'PRECEDES', 'SYMPTOM\_OF', 'PREDISPOSES'
    \item \textbf{BioRED} ~\cite{luo2022biored}: The Biological Relation Extraction Dataset, which includes detailed annotations of biological interactions within scientific texts.
    The relation types include 'Positive\_Correlation', 'Negative\_Correlation', 'Conversion', 'Drug\_Interaction', 'Cotreatment', 'Comparison', 'Association', 'Bind', totally 8 relation types.
\end{itemize}

For \textbf{Named Entity Recognition} task, our chosen datasets are:
\begin{itemize}
    \item \textbf{BC5CDR} ~\cite{li2016biocreative}: This dataset contains annotated mentions of chemicals and diseases in biomedical literature. Named entities include 'B-Chemical', 'B-Disease', 'I-Disease', 'I-Chemical' and 'else', which are mapped to 1,2,3,4,0 respectively in response.
    \item \textbf{BC2GM} ~\cite{smith2008overview}: The BioCreative II Gene Mention dataset, used for recognizing gene names. In response, gene names are mapped to 1 and others to 0.
    \item \textbf{BC4CHEMD} ~\cite{krallinger2015chemdner}: The BioCreative IV Chemical and Drug dataset, providing comprehensive annotations for chemical and drug names. The model need to map 'B-chemical' to 3, 'I-chemical' to 4, 'others' to 0.
\end{itemize}

For \textbf{Text Classification} task, we utilize:
\begin{itemize}
    \item \textbf{ADE} ~\cite{GURULINGAPPA2012885}: The Adverse Drug Events dataset could be used for binary classification task. The goal is to determine if the input sentence is ADE-related or not.
    \item \textbf{PubMed20krct} ~\cite{dernoncourt2017pubmed}: A large dataset of biomedical literature from PubMed. The task requires the model to classify the sentence into 5 categories: 'background', 'conclusions', 'methods', 'objective', 'results'.
    \item \textbf{HealthAdvice} ~\cite{yu-etal-2019-detecting}: A dataset containing health-related information and advice. The model need to classify the sentence to three categories: 'no advice', 'week advice', 'strong advice'.
\end{itemize}

\subsection{Model}
% Our framework could be used for any LLMs. Our primary focus is on exploring the potential of MTL. For our experiments, we selected the LLama3-8B \footnote{https://github.com/meta-llama/llama3} model to investigate its learning capabilities across multiple dataset settings. 

The specific model used in this paper is not the focus of our paper, as the proposed framework is designed to be applicable to any LLMs. Therefore, we opted for a representative open-source model: LLama3-8B \footnote{https://github.com/meta-llama/llama3}. 
It was selected due to its widespread use and strong performance across various NLP tasks, making it an ideal candidate for demonstrating the effectiveness of our dataset selection framework. Additionally, its open-source nature ensures reproducibility and allows for further experimentation by the research community.

\subsection{Data preparation}
To begin, it is necessary to generate several dataset combinations as basic data (including the single task learning as baseline), which will later allow the neural network to learn the relationships between datasets. 
For the 12 datasets we selected, we run combinations both within the same task and across different tasks. 
These combinations ensure that all datasets are involved, providing a diverse and comprehensive basis for understanding how different datasets interact and contribute to overall performance.

\subsection{Neural Network}
The choice of neural network architecture is flexible. In our experiment, we used a two-layer neural network with 12 inputs and 1 output to perform a regression task. The 12 inputs correspond to the 12 datasets: if a dataset is used, the corresponding input is set to 1; if it is not used, the input is set to 0. 
The output of the neural network is the score we aim to predict, such as the F1 score in our experiments. This design directly establishes a relationship between the use of specific datasets and the resulting F1 score.

Each time a new combination of datasets is trained and tested, the neural network is trained again. After training, we input all possible combinations and use the predicted F1 scores to determine the best combination for the next iteration and keep optimizing model performance.

\subsection{Instruction-tuning}

The prompt examples we used are shown in Fig. \ref{fig:3_prompt}.
The training prompt example includes instruction, an input sentence, and its corresponding response. By contrast, the testing prompt is empty in response, which allows the model to continue to generate the next tokens. $\backslash$n$\backslash$n is the ending mark for completing the response generation. Then in the evaluation stage, we will extract output tokens before the ending token.

Since the size of each dataset varies, to be fair, we instruction-tuned each model for 5000 steps, no matter how many tasks we included. We saved the model every 1000 steps and used the best model for the latter generation. When using multiple datasets, we sampled evenly from each one to make sure they were equally weighted.

\subsection{Metrics}

In the evaluation stage, we used Micro Precision, Recall, and F1-score.
When using these metrics, a prediction is considered correct only if the entire predicted output exactly matches the ground truth. 
% This strict criterion ensures that partial correctness for one input is not rewarded, reflecting the need for complete accuracy in our evaluations.
% For example, for NER task, if the input sentence is "['Effects', 'of', 'docosahexaenoic', 'acid', 'and', 'methylmercury', 'on', 'child', " ' ", 's', 'brain', 'development', 'due', 'to', 'consumption', 'of', 'fish', 'by', 'Finnish', 'mother', 'during', 'pregnancy', ':', 'a', 'probabilistic', 'modeling', 'approach', '.'], then the output should be exactly correct for every token as [0, 0, 3, 0, 4, 0, 3, 0, 0, 0, 0, 0, 0, 0, 0, 0, 0, 0, 0, 0, 0, 0, 0, 0, 0, 0, 0, 0, 0, 0], then it is counted as one correct answer.

%% file: 6_experiment.tex
\section{Results}
\subsection{Data preparation for RE task}
We included all attempts related to the RE task in Table \ref{table:RE_result}. As shown, the combinations involving BioRED, DDI demonstrate performance improvements for some specific combinations, while for the remaining GIT datasets, the baseline achieved the highest F1 score.
Specifically, for the BioRED 
\begin{table*}[h!]
\centering
\caption{Data preparation for RE task, including BioRED, DDI, and GIT datasets. $\bigstar$ indicates the best F1 score and best combination in the data preparation phase. Each line represents one experiment and $\surd$ means the dataset is selected for this run.}
\resizebox{\textwidth}{115mm}{
\begin{tabular}{ccccccccccccccccc}
\toprule
\multicolumn{12}{c}{Training sets}\\
\cmidrule(lr){1-12}
\multicolumn{3}{c}{RE task} & \multicolumn{3}{c}{NER task} & \multicolumn{3}{c}{EE task} & \multicolumn{3}{c}{TC task} & & & \multicolumn{3}{c}{Metrics} \\
\cmidrule(lr){1-3} \cmidrule(lr){4-6} \cmidrule(lr){7-9} \cmidrule(lr){10-12} \cmidrule(lr){15-17}
BioRED & DDI & GIT & BC2GM & BC4CHEMD & BC5CDR & GENIA2011 & GENIA2013 & PHEE & ADE & HealthAdvice & PubMed20krct & Task & Test Set & Precision & Recall & F1 Score \\
\midrule
$\surd$ &  &  &  &  &  &  &  &  &  &  &  & Baseline & BioRED & 37.11 & 95.95 & 53.52 \\
\hdashline
$\surd$ &  & $\surd$ &  &  &  &  &  &  &  &  &  & RE & BioRED & 36.23 & 97.75 &   52.86 $\downarrow$\\
$\surd$ & $\surd$ &  &  &  &  &  &  &  &  &  &  & RE & BioRED & 37.77 & 95.95 &   54.20 $\uparrow$\\
$\surd$ &  &  &  &  &  &  &  &  & $\surd$ &  &  & RE & BioRED & 37.45 & 91.44 &   53.14 $\downarrow$\\
$\surd$ &  &  &  &  &  & $\surd$ &  &  &  &  &  & RE & BioRED & 38.57 & 97.30 &   55.24 $\uparrow$\\
$\surd$ &  &  & $\surd$ &  &  &  &  &  &  &  &  & RE & BioRED & 40.65 & 95.95 &   57.10 $\uparrow$\\
$\surd$ & $\surd$ & $\surd$ &  &  &  &  &  &  &  &  &  & RE & BioRED & 40.22 & 97.30 &   56.92 $\uparrow$\\
$\surd$ &  &  & $\surd$ &  &  & $\surd$ &  &  &  &  &  & RE & BioRED & 37.99 & 96.85 &   54.57 $\uparrow$\\
$\surd$ &  &  &  &  &  & $\surd$ &  &  & $\surd$ &  &  & RE & BioRED & 46.84 & 90.09 &   61.63 $\uparrow$\\
$\surd$ &  &  & $\surd$ &  &  &  &  &  & $\surd$ &  &  & RE & BioRED & 41.62 & 97.30 &   58.30 $\uparrow$\\
$\surd$ &  &  & $\surd$ &  &  & $\surd$ &  &  & $\surd$ &  &  & RE & BioRED & 42.69 & 97.30 &   59.34 $\uparrow$\\
$\surd$ &  &  & $\surd$ &  &  & $\surd$ &  &  &  & $\surd$ &  & RE & BioRED & 43.93 & 94.59 &   60.00 $\uparrow$\\
$\surd$ &  &  & $\surd$ &  &  & $\surd$ &  &  &  &  & $\surd$ & RE & BioRED & 41.73 & 95.50 &   58.08 $\uparrow$\\
$\surd$ &  &  &  & $\surd$ &  & $\surd$ &  &  & $\surd$ &  &  & RE & BioRED & 42.15 & 95.50 &   58.48 $\uparrow$\\
$\surd$ &  &  &  & $\surd$ &  & $\surd$ &  &  &  & $\surd$ &  & RE & BioRED & 44.69 & 92.79 &   60.32 $\uparrow$\\
$\surd$ &  &  &  & $\surd$ &  & $\surd$ &  &  &  &  & $\surd$ & RE & BioRED & 41.15 & 96.40 &   57.68 $\uparrow$\\
$\surd$ &  &  & $\surd$ &  &  &  &  & $\surd$ & $\surd$ &  &  & RE & BioRED & 40.34 & 95.95 &   56.80 $\uparrow$\\
$\surd$ &  &  &  &  & $\surd$ & $\surd$ &  &  & $\surd$ &  &  & RE & BioRED & 42.97 & 97.75 &   59.70 $\uparrow$\\
$\surd$ &  &  &  &  & $\surd$ & $\surd$ &  &  &  &  & $\surd$ & RE & BioRED & 38.92 & 97.30 &   55.60 $\uparrow$\\
$\surd$ &  &  &  &  & $\surd$ & $\surd$ &  &  &  & $\surd$ &  & RE & BioRED & 48.18 & 95.50 &   64.05 $\bigstar$ \\
$\surd$ &  &  & $\surd$ &  &  &  &  & $\surd$ &  & $\surd$ &  & RE & BioRED & 40.26 & 96.85 &   56.88 $\uparrow$\\
$\surd$ &  &  &  & $\surd$ &  &  &  & $\surd$ & $\surd$ &  &  & RE & BioRED & 39.85 & 96.40 &   56.39 $\uparrow$\\
$\surd$ &  &  &  & $\surd$ &  &  &  & $\surd$ &  & $\surd$ &  & RE & BioRED & 35.16 & 98.20 &   51.78 $\downarrow$\\
$\surd$ &  &  &  & $\surd$ &  &  &  & $\surd$ &  &  & $\surd$ & RE & BioRED & 42.48 & 95.50 &   58.81 $\uparrow$\\
$\surd$ &  &  &  &  & $\surd$ &  &  & $\surd$ & $\surd$ &  &  & RE & BioRED & 39.63 & 96.40 &   56.17 $\uparrow$\\
$\surd$ &  &  & $\surd$ &  &  &  &  & $\surd$ &  &  & $\surd$ & RE & BioRED & 44.61 & 95.05 &   60.72 $\uparrow$\\
$\surd$ &  &  &  &  & $\surd$ &  &  & $\surd$ &  & $\surd$ &  & RE & BioRED & 40.80 & 96.85 &   57.41 $\uparrow$\\
$\surd$ &  &  &  &  & $\surd$ &  &  & $\surd$ &  &  & $\surd$ & RE & BioRED & 46.67 & 94.59 &   62.50 $\uparrow$\\
$\surd$ &  &  & $\surd$ &  &  &  & $\surd$ &  & $\surd$ &  &  & RE & BioRED & 44.57 & 86.94 &   58.93 $\uparrow$\\
$\surd$ &  &  & $\surd$ &  &  &  & $\surd$ &  &  & $\surd$ &  & RE & BioRED & 39.23 & 96.85 &   55.84 $\uparrow$\\
$\surd$ &  &  &  & $\surd$ &  &  & $\surd$ &  & $\surd$ &  &  & RE & BioRED & 37.16 & 91.89 &   52.92 $\downarrow$\\
$\surd$ &  &  &  & $\surd$ &  &  & $\surd$ &  &  & $\surd$ &  & RE & BioRED & 33.65 & 96.40 &   49.88 $\downarrow$\\
$\surd$ &  &  &  & $\surd$ &  &  & $\surd$ &  &  &  & $\surd$ & RE & BioRED & 39.17 & 93.69 &   55.25 $\uparrow$\\
$\surd$ &  &  &  &  & $\surd$ &  & $\surd$ &  & $\surd$ &  &  & RE & BioRED & 38.64 & 97.30 &   55.31 $\uparrow$\\
$\surd$ &  &  & $\surd$ &  &  &  & $\surd$ &  &  &  & $\surd$ & RE & BioRED & 38.60 & 96.85 &   55.20 $\uparrow$\\
$\surd$ &  &  &  &  & $\surd$ &  & $\surd$ &  &  & $\surd$ &  & RE & BioRED & 34.76 & 98.65 &   51.41 $\downarrow$\\
$\surd$ &  &  &  &  & $\surd$ &  & $\surd$ &  &  &  & $\surd$ & RE & BioRED & 39.17 & 97.75 &   55.93 $\uparrow$\\
\hdashline
 & $\surd$ &  &  &  &  &  &  &  &  &  &  & Baseline & DDI & 71.10 & 71.10 & 71.10 \\
 \hdashline
 & $\surd$ & $\surd$ &  &  &  &  &  &  &  &  &  & RE & DDI & 72.60 & 72.60 &   72.60 $\uparrow$\\
$\surd$ & $\surd$ &  &  &  &  &  &  &  &  &  &  & RE & DDI & 38.98 & 67.90 &   49.53 $\downarrow$\\
 & $\surd$ &  &  &  &  &  &  &  &  & $\surd$ &  & RE & DDI & 73.20 & 73.20 &   73.20 $\uparrow$\\
 & $\surd$ &  &  &  &  &  &  & $\surd$ &  &  &  & RE & DDI & 48.30 & 73.90 &   58.42 $\downarrow$\\
 & $\surd$ &  &  & $\surd$ &  &  &  &  &  &  &  & RE & DDI & 72.40 & 72.40 &   72.40 $\uparrow$\\
$\surd$ & $\surd$ & $\surd$ &  &  &  &  &  &  &  &  &  & RE & DDI & 64.21 & 65.30 &   64.75 $\downarrow$\\
 & $\surd$ &  &  & $\surd$ &  &  &  & $\surd$ &  &  &  & RE & DDI & 61.70 & 67.50 &   64.47 $\downarrow$\\
 & $\surd$ &  &  &  &  &  &  & $\surd$ &  & $\surd$ &  & RE & DDI & 61.52 & 68.10 &   64.64 $\downarrow$\\
 & $\surd$ &  &  & $\surd$ &  &  &  &  &  & $\surd$ &  & RE & DDI & 73.68 & 74.20 &   73.94 $\bigstar$ \\
 & $\surd$ &  &  & $\surd$ &  &  &  & $\surd$ &  & $\surd$ &  & RE & DDI & 68.92 & 69.20 &   69.06 $\downarrow$\\
 & $\surd$ &  & $\surd$ &  &  & $\surd$ &  &  & $\surd$ &  &  & RE & DDI & 31.31 & 76.30 &   44.40 $\downarrow$\\
 & $\surd$ &  & $\surd$ &  &  & $\surd$ &  &  &  & $\surd$ &  & RE & DDI & 37.86 & 71.90 &   49.60 $\downarrow$\\
 & $\surd$ &  & $\surd$ &  &  & $\surd$ &  &  &  &  & $\surd$ & RE & DDI & 54.99 & 68.90 &   61.16 $\downarrow$\\
 & $\surd$ &  &  & $\surd$ &  & $\surd$ &  &  & $\surd$ &  &  & RE & DDI & 30.20 & 66.60 &   41.56 $\downarrow$\\
 & $\surd$ &  &  & $\surd$ &  & $\surd$ &  &  &  & $\surd$ &  & RE & DDI & 37.97 & 73.70 &   50.12 $\downarrow$\\
 & $\surd$ &  &  & $\surd$ &  & $\surd$ &  &  &  &  & $\surd$ & RE & DDI & 60.56 & 68.80 &   64.42 $\downarrow$\\
 & $\surd$ &  &  &  & $\surd$ & $\surd$ &  &  & $\surd$ &  &  & RE & DDI & 34.35 & 80.20 &   48.10 $\downarrow$\\
 & $\surd$ &  &  &  & $\surd$ & $\surd$ &  &  &  &  & $\surd$ & RE & DDI & 67.07 & 67.20 &   67.13 $\downarrow$\\
 & $\surd$ &  &  &  & $\surd$ & $\surd$ &  &  &  & $\surd$ &  & RE & DDI & 53.15 & 65.00 &   58.48 $\downarrow$\\
 & $\surd$ &  & $\surd$ &  &  &  &  & $\surd$ & $\surd$ &  &  & RE & DDI & 65.68 & 70.80 &   68.14 $\downarrow$\\
 & $\surd$ &  & $\surd$ &  &  &  &  & $\surd$ &  & $\surd$ &  & RE & DDI & 69.80 & 69.80 &   69.80 $\downarrow$\\
 & $\surd$ &  & $\surd$ &  &  &  &  & $\surd$ &  &  & $\surd$ & RE & DDI & 67.80 & 67.80 &   67.80 $\downarrow$\\
 & $\surd$ &  &  &  & $\surd$ &  &  & $\surd$ & $\surd$ &  &  & RE & DDI & 61.88 & 72.40 &   66.73 $\downarrow$\\
 & $\surd$ &  &  & $\surd$ &  &  &  & $\surd$ &  &  & $\surd$ & RE & DDI & 64.29 & 64.80 &   64.54 $\downarrow$\\
 & $\surd$ &  &  &  & $\surd$ &  &  & $\surd$ &  & $\surd$ &  & RE & DDI & 61.95 & 65.30 &   63.58 $\downarrow$\\
 & $\surd$ &  &  &  & $\surd$ &  &  & $\surd$ &  &  & $\surd$ & RE & DDI & 62.41 & 68.90 &   65.49 $\downarrow$\\
 & $\surd$ &  & $\surd$ &  &  &  & $\surd$ &  & $\surd$ &  &  & RE & DDI & 29.27 & 53.30 &   37.79 $\downarrow$\\
 & $\surd$ &  & $\surd$ &  &  &  & $\surd$ &  &  & $\surd$ &  & RE & DDI & 29.93 & 78.20 &   43.29 $\downarrow$\\
 & $\surd$ &  & $\surd$ &  &  &  & $\surd$ &  &  &  & $\surd$ & RE & DDI & 62.13 & 67.60 &   64.75 $\downarrow$\\
 & $\surd$ &  &  & $\surd$ &  &  & $\surd$ &  & $\surd$ &  &  & RE & DDI & 25.19 & 63.80 &   36.12 $\downarrow$\\
 & $\surd$ &  &  & $\surd$ &  &  & $\surd$ &  &  & $\surd$ &  & RE & DDI & 21.57 & 60.30 &   31.78 $\downarrow$\\
 & $\surd$ &  &  & $\surd$ &  &  & $\surd$ &  &  &  & $\surd$ & RE & DDI & 62.65 & 66.10 &   64.33 $\downarrow$\\
 & $\surd$ &  &  &  & $\surd$ &  & $\surd$ &  & $\surd$ &  &  & RE & DDI & 60.96 & 68.70 &   64.60 $\downarrow$\\
 & $\surd$ &  &  &  & $\surd$ &  & $\surd$ &  &  & $\surd$ &  & RE & DDI & 36.25 & 70.80 &   47.95 $\downarrow$\\
 & $\surd$ &  &  &  & $\surd$ &  & $\surd$ &  &  &  & $\surd$ & RE & DDI & 66.32 & 70.50 &   68.35 $\downarrow$\\
 \hdashline
 &  & $\surd$ &  &  &  &  &  &  &  &  &  & Baseline & GIT & 77.20 & 77.20 & 77.20 $\bigstar$ \\
 \hdashline
$\surd$ &  & $\surd$ &  &  &  &  &  &  &  &  &  & RE & GIT & 17.55 & 80.65 &   28.82 $\downarrow$\\
 & $\surd$ & $\surd$ &  &  &  &  &  &  &  &  &  & RE & GIT & 64.52 & 64.52 &   64.52 $\downarrow$\\
 &  & $\surd$ &  &  & $\surd$ &  &  &  &  &  &  & RE & GIT & 67.31 & 67.31 &   67.31 $\downarrow$\\
 &  & $\surd$ &  &  &  &  &  &  &  &  & $\surd$ & RE & GIT & 66.67 & 66.67 &   66.67 $\downarrow$\\
 &  & $\surd$ &  &  &  &  & $\surd$ &  &  &  &  & RE & GIT & 43.65 & 67.31 &   52.96 $\downarrow$\\
$\surd$ & $\surd$ & $\surd$ &  &  &  &  &  &  &  &  &  & RE & GIT & 15.97 & 70.54 &   26.04 $\downarrow$\\
 &  & $\surd$ &  &  & $\surd$ &  & $\surd$ &  &  &  &  & RE & GIT & 50.75 & 58.49 &   54.35 $\downarrow$\\
 &  & $\surd$ &  &  &  &  & $\surd$ &  &  &  & $\surd$ & RE & GIT & 20.42 & 71.83 &   31.79 $\downarrow$\\
 &  & $\surd$ &  &  & $\surd$ &  &  &  &  &  & $\surd$ & RE & GIT & 57.20 & 57.20 &   57.20 $\downarrow$\\
 &  & $\surd$ &  &  & $\surd$ &  & $\surd$ &  &  &  & $\surd$ & RE & GIT & 46.21 & 58.92 &   51.80 $\downarrow$\\
 &  & $\surd$ &  &  & $\surd$ &  &  & $\surd$ &  & $\surd$ &  & RE & GIT & 41.87 & 47.10 &   44.33 $\downarrow$\\
 &  & $\surd$ &  &  & $\surd$ &  &  & $\surd$ &  &  & $\surd$ & RE & GIT & 28.45 & 43.01 &   34.25 $\downarrow$\\
 &  & $\surd$ & $\surd$ &  &  &  & $\surd$ &  & $\surd$ &  &  & RE & GIT & 25.73 & 66.67 &   37.13 $\downarrow$\\
 &  & $\surd$ & $\surd$ &  &  &  & $\surd$ &  &  & $\surd$ &  & RE & GIT & 17.96 & 64.30 &   28.08 $\downarrow$\\
 &  & $\surd$ & $\surd$ &  &  &  & $\surd$ &  &  &  & $\surd$ & RE & GIT & 43.04 & 58.49 &   49.59 $\downarrow$\\
 &  & $\surd$ &  & $\surd$ &  &  & $\surd$ &  & $\surd$ &  &  & RE & GIT & 18.54 & 73.12 &   29.58 $\downarrow$\\
 &  & $\surd$ &  & $\surd$ &  &  & $\surd$ &  &  & $\surd$ &  & RE & GIT & 26.88 & 60.86 &   37.29 $\downarrow$\\
 &  & $\surd$ &  & $\surd$ &  &  & $\surd$ &  &  &  & $\surd$ & RE & GIT & 31.86 & 54.41 &   40.19 $\downarrow$\\
 &  & $\surd$ &  &  & $\surd$ &  & $\surd$ &  & $\surd$ &  &  & RE & GIT & 17.37 & 70.97 &   27.91 $\downarrow$\\
 &  & $\surd$ &  &  & $\surd$ &  & $\surd$ &  &  & $\surd$ &  & RE & GIT & 27.68 & 64.95 &   38.82 $\downarrow$\\
 &  & $\surd$ & $\surd$ &  &  & $\surd$ &  &  &  &  & $\surd$ & RE & GIT & 41.55 & 59.78 &   49.03 $\downarrow$\\
 &  & $\surd$ &  & $\surd$ &  & $\surd$ &  &  &  & $\surd$ &  & RE & GIT & 16.04 & 76.56 &   26.52 $\downarrow$\\
 &  & $\surd$ &  & $\surd$ &  & $\surd$ &  &  & $\surd$ &  &  & RE & GIT & 16.73 & 73.76 &   27.28 $\downarrow$\\
 &  & $\surd$ &  & $\surd$ &  & $\surd$ &  &  &  &  & $\surd$ & RE & GIT & 18.71 & 68.39 &   29.38 $\downarrow$\\
 &  & $\surd$ &  &  & $\surd$ & $\surd$ &  &  & $\surd$ &  &  & RE & GIT & 35.96 & 62.80 &   45.73 $\downarrow$\\
 &  & $\surd$ &  &  & $\surd$ & $\surd$ &  &  &  & $\surd$ &  & RE & GIT & 20.28 & 67.96 &   31.24 $\downarrow$\\
 &  & $\surd$ &  &  & $\surd$ & $\surd$ &  &  &  &  & $\surd$ & RE & GIT & 33.10 & 61.51 &   43.04 $\downarrow$\\
 &  & $\surd$ & $\surd$ &  &  &  &  & $\surd$ & $\surd$ &  &  & RE & GIT & 47.53 & 47.53 &   47.53 $\downarrow$\\
 &  & $\surd$ & $\surd$ &  &  &  &  & $\surd$ &  & $\surd$ &  & RE & GIT & 34.60 & 54.62 &   42.37 $\downarrow$\\
 &  & $\surd$ & $\surd$ &  &  &  &  & $\surd$ &  &  & $\surd$ & RE & GIT & 23.22 & 60.43 &   33.55 $\downarrow$\\
 &  & $\surd$ &  & $\surd$ &  &  &  & $\surd$ & $\surd$ &  &  & RE & GIT & 17.65 & 31.40 &   22.60 $\downarrow$\\
 &  & $\surd$ &  & $\surd$ &  &  &  & $\surd$ &  & $\surd$ &  & RE & GIT & 29.33 & 60.86 &   39.58 $\downarrow$\\
 &  & $\surd$ &  & $\surd$ &  &  &  & $\surd$ &  &  & $\surd$ & RE & GIT & 19.28 & 46.02 &   27.17 $\downarrow$\\
 &  & $\surd$ &  &  & $\surd$ &  &  & $\surd$ & $\surd$ &  &  & RE & GIT & 17.93 & 64.52 &   28.06 $\downarrow$\\
 &  & $\surd$ & $\surd$ &  &  & $\surd$ &  &  & $\surd$ &  &  & RE & GIT & 18.51 & 66.45 &   28.96 $\downarrow$\\
 &  & $\surd$ & $\surd$ &  &  & $\surd$ &  &  &  & $\surd$ &  & RE & GIT & 28.73 & 54.19 &   37.56 $\downarrow$\\
\bottomrule
\end{tabular}}
\label{table:RE_result}
\end{table*}
% \end{center}
% \vspace{-10.0em}

\begin{figure*}[htb!]
\begin{subfigure}{0.33\textwidth}
\includegraphics[width=0.9\linewidth]{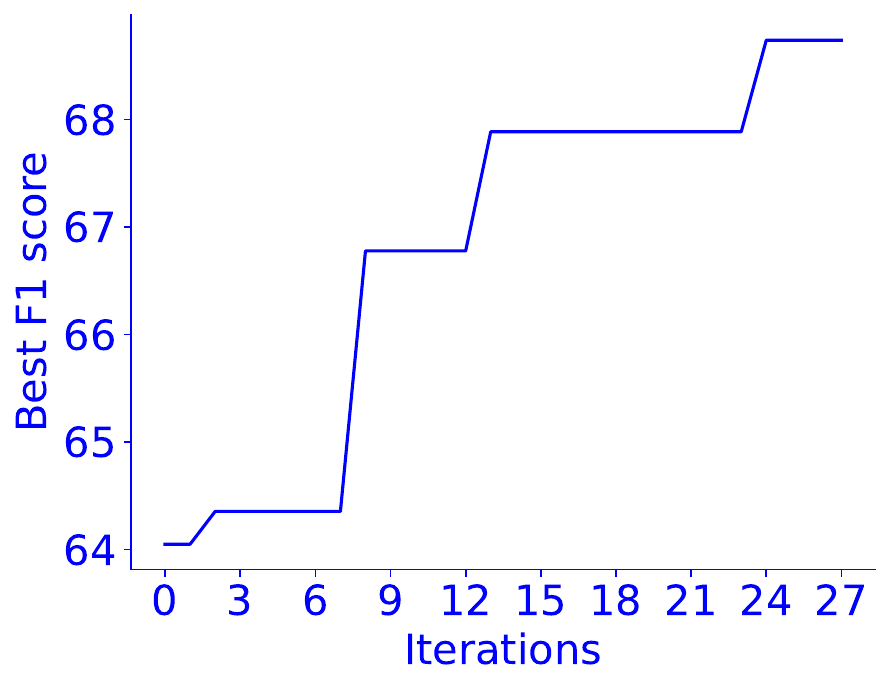}
\caption{BioRED}
\label{fig:0_stl}
\end{subfigure}
\begin{subfigure}{0.33\textwidth}
\includegraphics[width=0.9\linewidth]{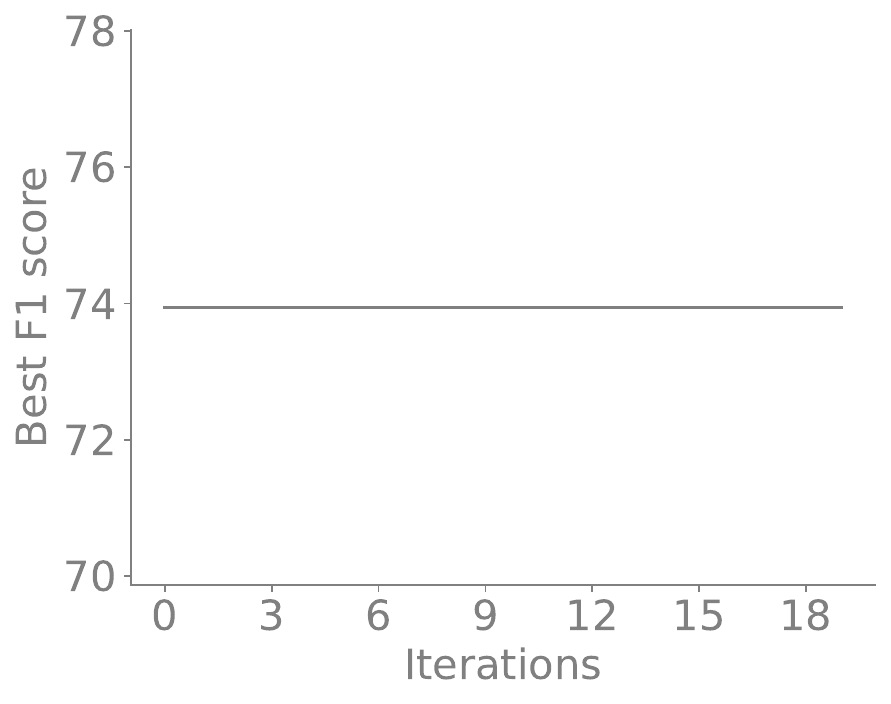}
\caption{DDI}
\label{fig:1_mtl}
\end{subfigure}
\begin{subfigure}{0.33\textwidth}
\includegraphics[width=0.9\linewidth]{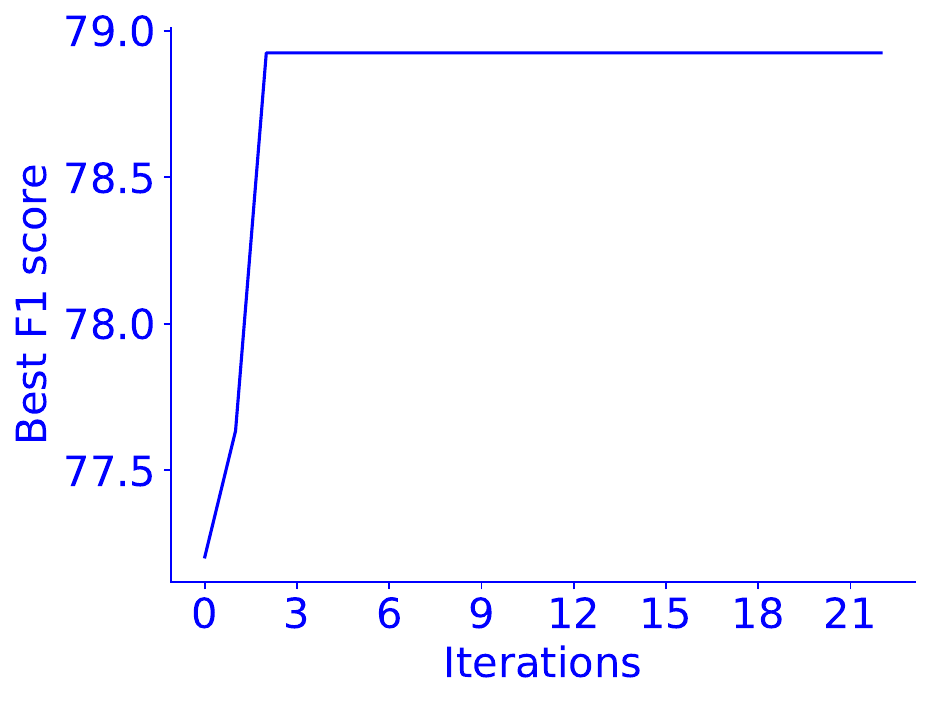}
\caption{GIT}
\label{fig:1_mtl}
\end{subfigure}
\caption{Framework results for datasets in the RE task.}
\label{fig:fw_RE}
\end{figure*}

\noindent 
dataset, 30/36 combinations show improvement, 4 combinations for DDI dataset demonstrate improvement, and all combinations we tried for the GIT dataset failed to bring in MTL gain.

As shown in Table. \ref{table:RE_result}, based on the results, the human experience might conclude that MTL works particularly well for the BioRED dataset, where the knowledge learned from other datasets aids the LLM in extracting relations from BioRED. For the DDI dataset, MTL offers slight improvements, but it does not provide any benefits for the GIT dataset.

Another important observation from our data preparation experiments (also considering the results from the ablation study) is that for tasks with poor baseline performance (e.g., F1 scores below 60 in this paper), incorporating multiple related training datasets within our framework significantly improves the results. This suggests that for tasks that are more challenging for LLMs, MTL can effectively leverage diverse data sources to enhance learning and improve performance. In contrast, for tasks with already high baseline performance, the benefits of MTL are minimal, indicating that MTL may not provide substantial improvements for tasks that are already well-optimized. This finding highlights that the impact of MTL is influenced by the LLM's capability to handle the task, showcasing its potential in improving low-performing tasks, while offering limited gains for tasks that are already performing well.

\subsection{Find better combination for datasets in the RE task iteratively using the proposed framework}

We applied the framework to three datasets in the RE task separately. The framework was set to automatically run for 48 hours, exploring better dataset combinations and stopping once sufficient exploration was achieved, keeping only the best model and the highest F1 score.

When using the framework, for each dataset, after each iteration, we enumerate $2^{11}=2048$ kinds of combinations for the nerual network and find the best combination. 
The best F1 score curves for three datasets are shown in Fig. \ref{fig:fw_RE}.
For the BioRED dataset in Fig. \ref{fig:fw_RE}(a), our framework could effectively find better combinations for each of several iterations. 
In addition, the framework also helps improve the performance of the GIT dataset, and then it stops after some exploration, which is out of expectation and demonstrates the robustness of our framework because it can find better combinations even if our initial combinations are bad.
For the DDI dataset, the framework cannot find better combinations and stops soon, which matches our expectations because most of the cases show performance degradation.

\begin{figure*}[!h]
% \begin{subfigure}{0.33\textwidth}
% \includegraphics[width=0.9\linewidth]{biored.pdf}
% \caption{BioRED}
% \label{fig:0_stl}
% \end{subfigure}
% \begin{subfigure}{0.33\textwidth}
% \includegraphics[width=0.9\linewidth]{DDI.pdf}
% \caption{DDI}
% \label{fig:1_mtl}
% \end{subfigure}
% \begin{subfigure}{0.33\textwidth}
% \includegraphics[width=0.9\linewidth]{git.pdf}
% \caption{GIT}
% \label{fig:1_mtl}
% \end{subfigure}
\begin{subfigure}{0.33\textwidth}
\includegraphics[width=0.9\linewidth]{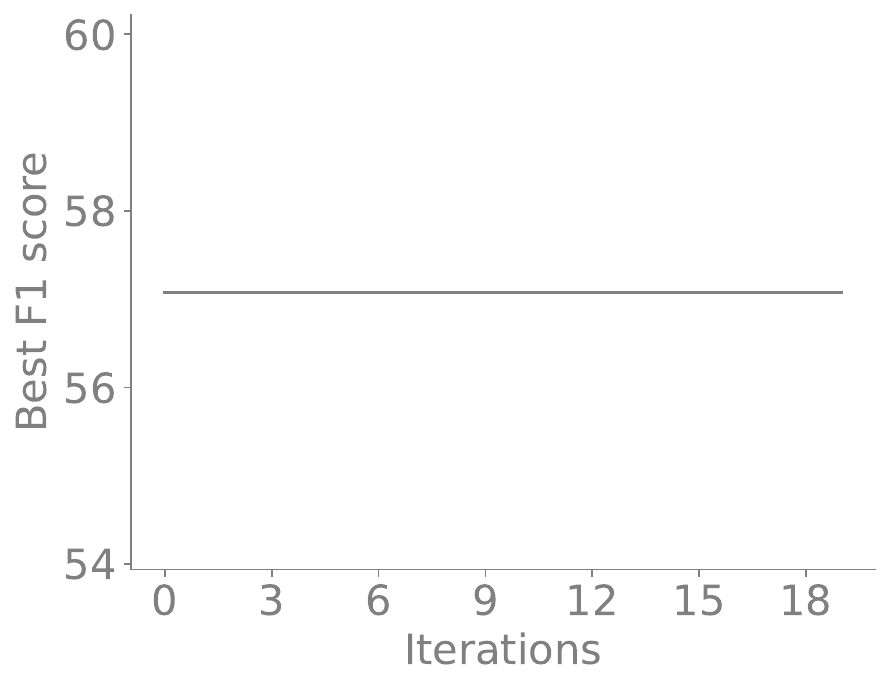}
\caption{Genia2011}
\label{fig:0_stl}
\end{subfigure}
\begin{subfigure}{0.33\textwidth}
\includegraphics[width=0.9\linewidth]{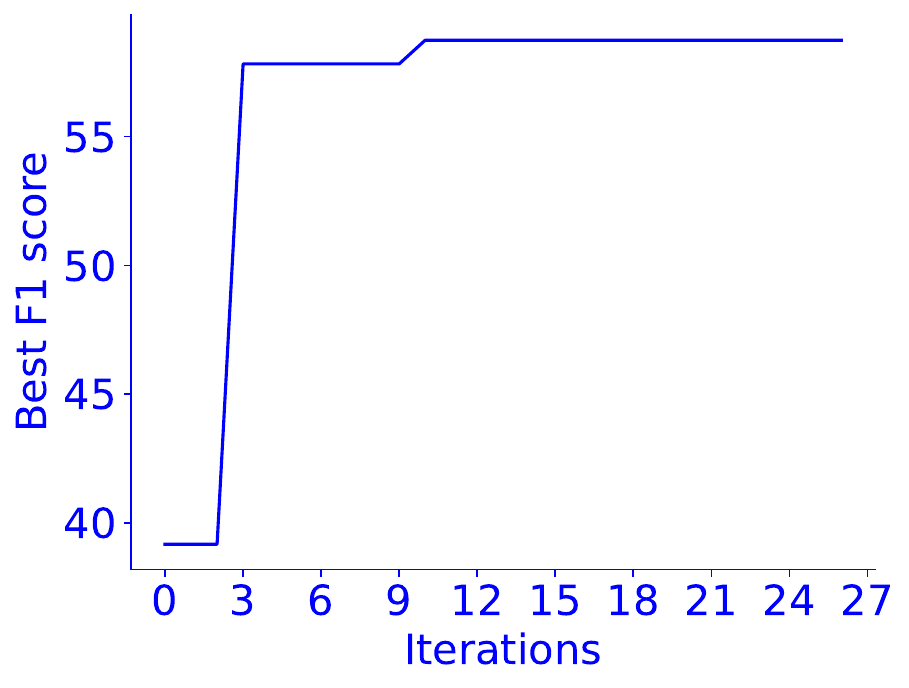}
\caption{Genia2013}
\label{fig:1_mtl}
\end{subfigure}
\begin{subfigure}{0.33\textwidth}
\includegraphics[width=0.9\linewidth]{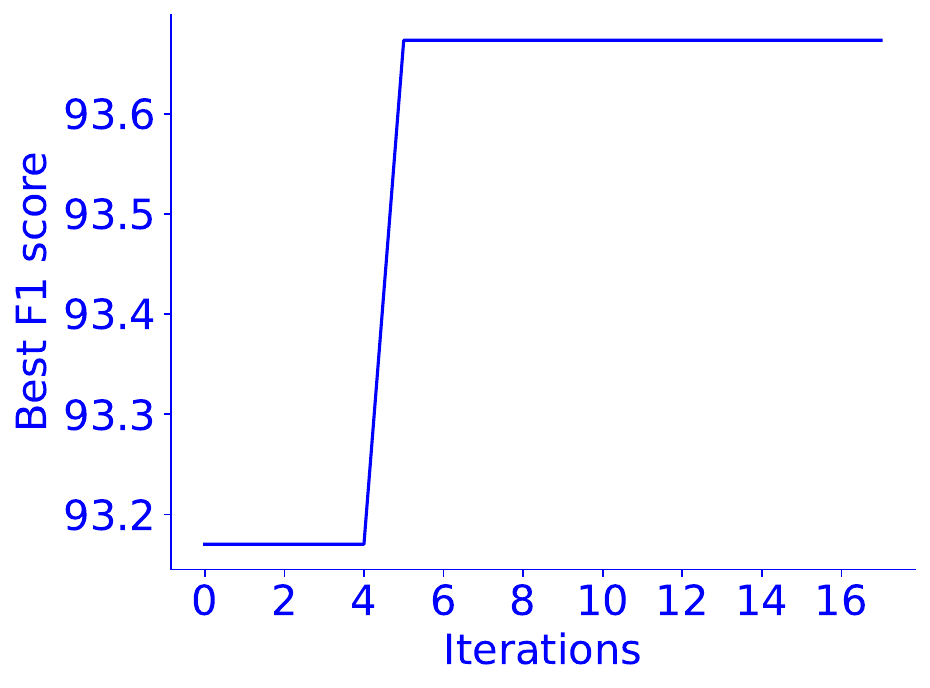}
\caption{PHEE}
\label{fig:1_mtl}
\end{subfigure}
\begin{subfigure}{0.33\textwidth}
\includegraphics[width=0.9\linewidth]{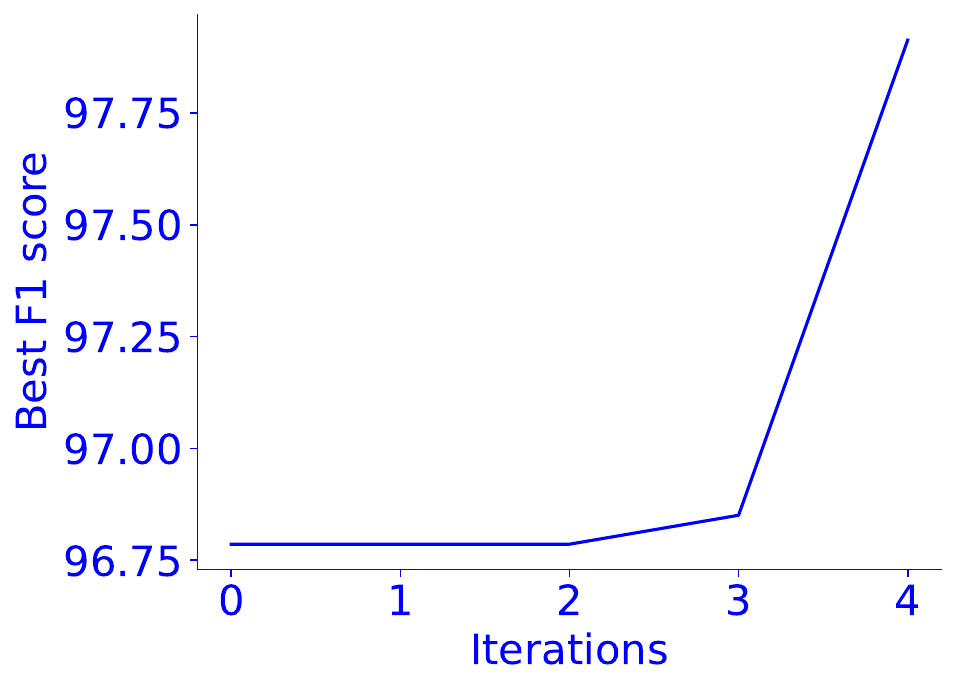}
\caption{BC2GM}
\label{fig:0_stl}
\end{subfigure}
\begin{subfigure}{0.33\textwidth}
\includegraphics[width=0.9\linewidth]{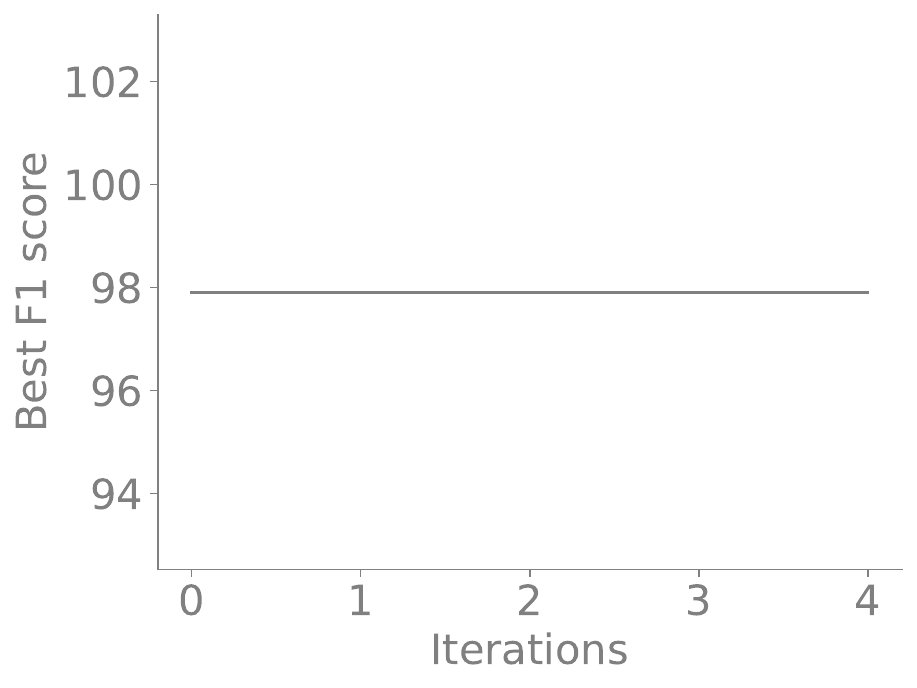}
\caption{BC4CHEMD}
\label{fig:1_mtl}
\end{subfigure}
\begin{subfigure}{0.33\textwidth}
\includegraphics[width=0.9\linewidth]{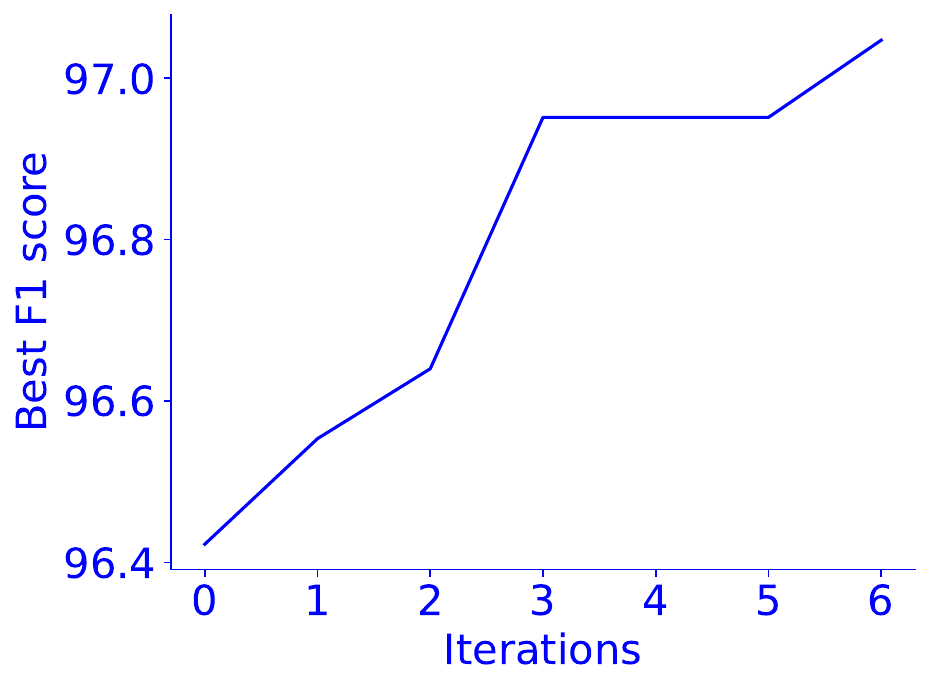}
\caption{BC5CDR}
\label{fig:1_mtl}
\end{subfigure}
\begin{subfigure}{0.33\textwidth}
\includegraphics[width=0.9\linewidth]{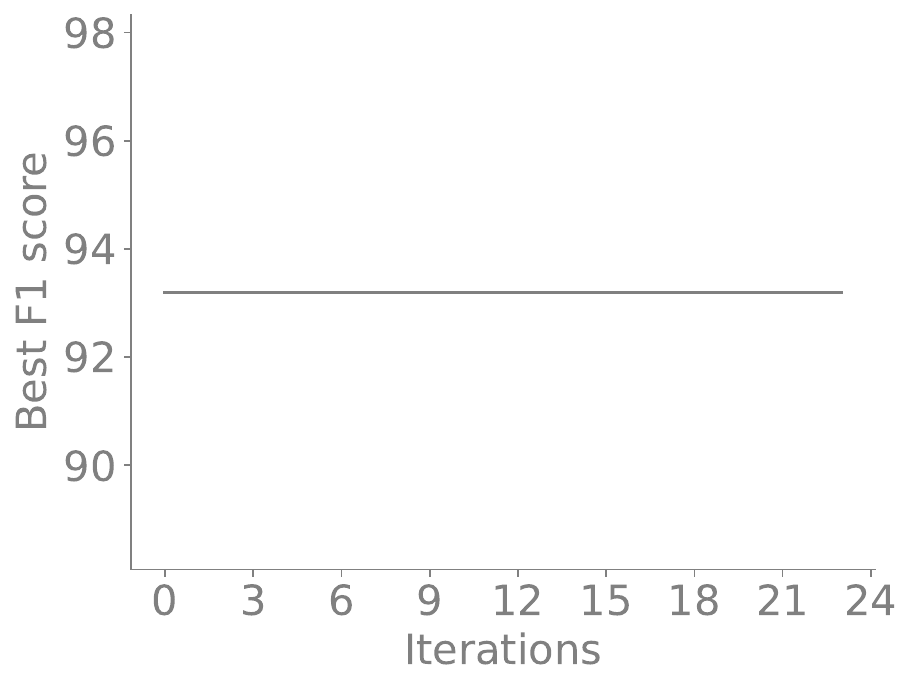}
\caption{ADE}
\label{fig:0_stl}
\end{subfigure}
\begin{subfigure}{0.33\textwidth}
\includegraphics[width=0.9\linewidth]{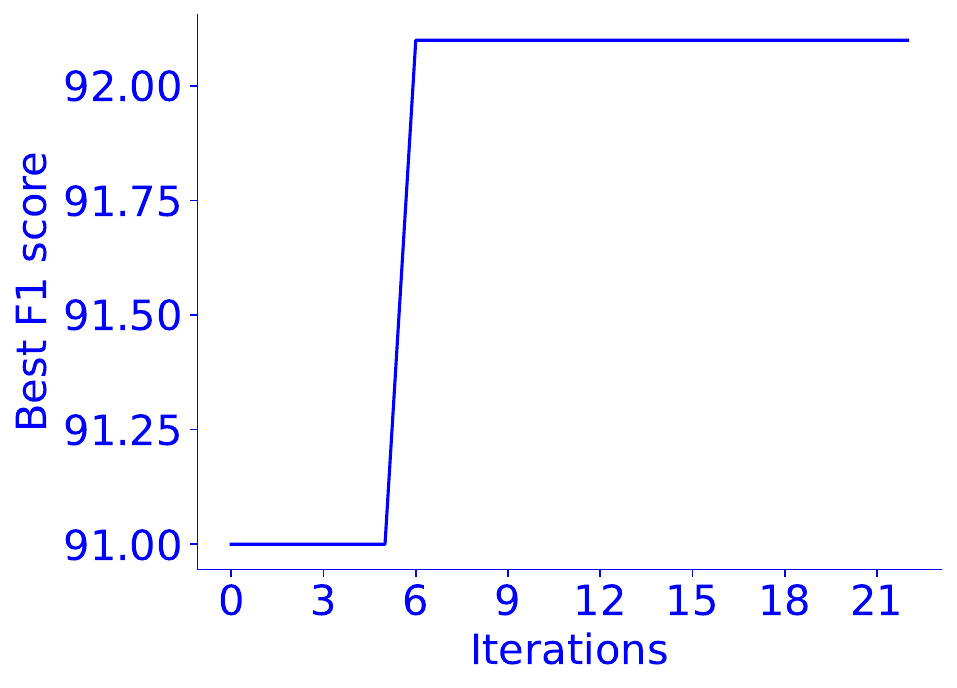}
\caption{HealthAdvice}
\label{fig:1_mtl}
\end{subfigure}
\begin{subfigure}{0.33\textwidth}
\includegraphics[width=0.9\linewidth]{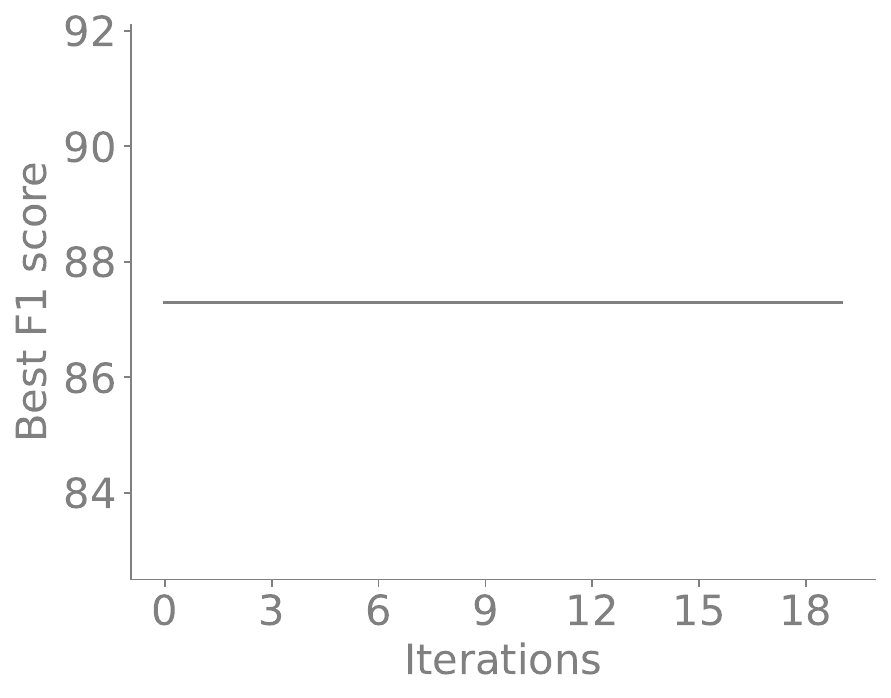}
\caption{PubMed20krct}
\label{fig:1_mtl}
\end{subfigure}
\caption{Finding best F1 results using proposed framework for EE (GENIA2011, GENIA2013, PHEE), NER (BC2GM, BC4CHEMD, BC5CDR), TC tasks (ADE, HealthAdvice, PubMed20krct).}
\end{figure*}

\subsection{Ablation study for other tasks}
Due to space limitation, the different combinations involving the other 9 datasets for NER, EE, and TC tasks are presented in the appendix, in Tables \ref{table:NER_result}, \ref{table:EE_result}, and \ref{table:TC_result}, respectively.

For the EE task, during the data preparation phase, we observed that several dataset combinations could improve LLM performance on GENIA2011 and GENIA2013, further demonstrating the potential of MTL. After applying our framework, the performance of GENIA2013 improved significantly from below 40 to around 58, a substantial gain. This result is corroborated by the improvements in GENIA2011, where an effective combination was tested early in the data preparation stage, boosting performance from a baseline of 34 to 57. Given the high similarity between GENIA2011 and GENIA2013, this suggests that if performance can be enhanced in GENIA2011, similar improvements should likely be achieved for GENIA2013. Although the initial combinations differed, our framework was able to find similar combinations for both datasets, ultimately achieving comparable results.

For the NER and TC tasks, the six datasets involved all achieved their best F1 scores with the baseline settings, and every combination we attempted resulted in a performance decline. From an expert perspective, it appeared that MTL could not leverage the other datasets in this paper to improve these six datasets. However, we continued to use our framework to investigate whether a specific combination could improve these seemingly unpromising datasets. Surprisingly, as shown in Figures 5(d), 5(f), and 5(h), our framework identified promising combinations that led to significant improvements for three of the datasets, despite initial expectations to the contrary. For the remaining three datasets (BC4CHEMD, ADE, PubMed20krct), we were unable to find better combinations, which aligns with our initial expectations.

Summarizing all the experiments, each dataset had 2,048 possible auxiliary combinations. However, in most cases, we were able to find relatively optimal combinations within just a dozen iterations, and the framework predicted that no better combinations existed. Compared to a brute-force approach, our method significantly improved efficiency.

%% file: 7_conclution.tex
\section{Conclusion}

We proposed a novel yet simple framework to address the challenge of selecting the optimal dataset combination for multi-task learning. 
% Our framework utilizes a simple neural network to predict the most promising dataset combination. 
By iteratively refining these combinations within a feedback loop, we take a significant step toward fully unlocking the potential of MTL in the future.

\section{Limitations}
The experiments in this paper were conducted using only a single LLM. Although the authors intended to experiment with multiple LLMs to explore broader performance variations, the computational resources and time required were prohibitive. As a result, the findings may not fully represent the potential of our framework.

Secondly, we did not perform a grid search to find the optimal hyperparameters for model training. 
Instead, we ensured that all experiments were conducted with the same set of parameters to maintain fairness.

%% file: 8_appendix.tex
\appendix

\section{Appendix}
\label{sec:appendix}

\begin{center}
\begin{table*}[t]
\centering
\caption{Data preparation for the NER task, including BC2GM, BC4CHEMD, and BC5CDR datasets. $\bigstar$ indicates the best F1 score and best combination in the data preparation phase. Each line represents one experiment and $\surd$ means the dataset is selected for this run.}
\resizebox{\textwidth}{110mm}{
\begin{tabular}{ccccccccccccccccc}
\toprule
\multicolumn{12}{c}{Training sets}\\
\cmidrule(lr){1-12}
\multicolumn{3}{c}{RE task} & \multicolumn{3}{c}{NER task} & \multicolumn{3}{c}{EE task} & \multicolumn{3}{c}{TC task} & & & \multicolumn{3}{c}{Metrics} \\
\cmidrule(lr){1-3} \cmidrule(lr){4-6} \cmidrule(lr){7-9} \cmidrule(lr){10-12} \cmidrule(lr){15-17}
BioRED & DDI & GIT & BC2GM & BC4CHEMD & BC5CDR & GENIA2011 & GENIA2013 & PHEE & ADE & HealthAdvice & PubMed20krct & Task & Test Set & Precision & Recall & F1 Score \\
\midrule
 &  &  & $\surd$ &  &  &  &  &  &  &  &  & Baseline & BC2GM & 98.01 & 95.60 & 96.78 $\bigstar$ \\
\hdashline
 &  &  & $\surd$ & $\surd$ &  &  &  &  &  &  &  & NER & BC2GM & 96.85 & 95.07 &   95.95 $\downarrow$\\
 &  &  & $\surd$ &  & $\surd$ &  &  &  &  &  &  & NER & BC2GM & 97.42 & 94.36 &   95.87 $\downarrow$\\
 &  &  & $\surd$ &  &  &  &  &  & $\surd$ &  &  & NER & BC2GM & 97.50 & 91.92 &   94.63 $\downarrow$\\
$\surd$ &  &  & $\surd$ &  &  &  &  &  &  &  &  & NER & BC2GM & 96.30 & 93.32 &   94.79 $\downarrow$\\
 &  &  & $\surd$ &  &  & $\surd$ &  &  &  &  &  & NER & BC2GM & 97.51 & 91.61 &   94.47 $\downarrow$\\
 &  &  & $\surd$ & $\surd$ & $\surd$ &  &  &  &  &  &  & NER & BC2GM & 96.76 & 92.77 &   94.72 $\downarrow$\\
$\surd$ &  &  & $\surd$ &  &  & $\surd$ &  &  &  &  &  & NER & BC2GM & 97.01 & 90.37 &   93.57 $\downarrow$\\
$\surd$ &  &  & $\surd$ &  &  &  &  &  & $\surd$ &  &  & NER & BC2GM & 96.78 & 91.05 &   93.83 $\downarrow$\\
 &  &  & $\surd$ &  &  & $\surd$ &  &  & $\surd$ &  &  & NER & BC2GM & 96.62 & 92.53 &   94.54 $\downarrow$\\
$\surd$ &  &  & $\surd$ &  &  & $\surd$ &  &  & $\surd$ &  &  & NER & BC2GM & 96.95 & 89.49 &   93.07 $\downarrow$\\
$\surd$ &  &  & $\surd$ &  &  & $\surd$ &  &  &  &  & $\surd$ & NER & BC2GM & 95.59 & 89.06 &   92.21 $\downarrow$\\
$\surd$ &  &  & $\surd$ &  &  & $\surd$ &  &  &  & $\surd$ &  & NER & BC2GM & 95.31 & 91.61 &   93.42 $\downarrow$\\
$\surd$ &  &  & $\surd$ &  &  &  &  & $\surd$ & $\surd$ &  &  & NER & BC2GM & 96.35 & 90.16 &   93.15 $\downarrow$\\
$\surd$ &  &  & $\surd$ &  &  &  &  & $\surd$ &  & $\surd$ &  & NER & BC2GM & 96.21 & 87.73 &   91.77 $\downarrow$\\
$\surd$ &  &  & $\surd$ &  &  &  &  & $\surd$ &  &  & $\surd$ & NER & BC2GM & 97.28 & 87.84 &   92.32 $\downarrow$\\
$\surd$ &  &  & $\surd$ &  &  &  & $\surd$ &  & $\surd$ &  &  & NER & BC2GM & 96.40 & 88.17 &   92.10 $\downarrow$\\
$\surd$ &  &  & $\surd$ &  &  &  & $\surd$ &  &  & $\surd$ &  & NER & BC2GM & 96.68 & 87.17 &   91.68 $\downarrow$\\
$\surd$ &  &  & $\surd$ &  &  &  & $\surd$ &  &  &  & $\surd$ & NER & BC2GM & 96.11 & 88.34 &   92.06 $\downarrow$\\
 & $\surd$ &  & $\surd$ &  &  & $\surd$ &  &  &  &  & $\surd$ & NER & BC2GM & 95.60 & 85.91 &   90.49 $\downarrow$\\
 & $\surd$ &  & $\surd$ &  &  & $\surd$ &  &  &  & $\surd$ &  & NER & BC2GM & 95.55 & 85.83 &   90.43 $\downarrow$\\
 & $\surd$ &  & $\surd$ &  &  & $\surd$ &  &  & $\surd$ &  &  & NER & BC2GM & 96.10 & 89.45 &   92.66 $\downarrow$\\
 & $\surd$ &  & $\surd$ &  &  &  &  & $\surd$ & $\surd$ &  &  & NER & BC2GM & 96.73 & 87.41 &   91.83 $\downarrow$\\
 &  & $\surd$ & $\surd$ &  &  &  & $\surd$ &  &  & $\surd$ &  & NER & BC2GM & 96.76 & 89.36 &   92.91 $\downarrow$\\
 &  & $\surd$ & $\surd$ &  &  &  & $\surd$ &  & $\surd$ &  &  & NER & BC2GM & 96.52 & 90.03 &   93.16 $\downarrow$\\
 &  & $\surd$ & $\surd$ &  &  &  & $\surd$ &  &  &  & $\surd$ & NER & BC2GM & 96.12 & 89.27 &   92.57 $\downarrow$\\
 & $\surd$ &  & $\surd$ &  &  &  &  & $\surd$ &  & $\surd$ &  & NER & BC2GM & 96.71 & 87.45 &   91.85 $\downarrow$\\
 & $\surd$ &  & $\surd$ &  &  &  &  & $\surd$ &  &  & $\surd$ & NER & BC2GM & 96.66 & 87.42 &   91.81 $\downarrow$\\
 & $\surd$ &  & $\surd$ &  &  &  & $\surd$ &  &  &  & $\surd$ & NER & BC2GM & 96.26 & 89.05 &   92.51 $\downarrow$\\
 & $\surd$ &  & $\surd$ &  &  &  & $\surd$ &  & $\surd$ &  &  & NER & BC2GM & 95.45 & 89.64 &   92.45 $\downarrow$\\
 & $\surd$ &  & $\surd$ &  &  &  & $\surd$ &  &  & $\surd$ &  & NER & BC2GM & 96.52 & 87.05 &   91.54 $\downarrow$\\
 &  & $\surd$ & $\surd$ &  &  & $\surd$ &  &  &  &  & $\surd$ & NER & BC2GM & 96.86 & 87.59 &   91.99 $\downarrow$\\
 &  & $\surd$ & $\surd$ &  &  &  &  & $\surd$ &  & $\surd$ &  & NER & BC2GM & 95.95 & 92.47 &   94.18 $\downarrow$\\
 &  & $\surd$ & $\surd$ &  &  &  &  & $\surd$ &  &  & $\surd$ & NER & BC2GM & 96.79 & 90.17 &   93.36 $\downarrow$\\
 &  & $\surd$ & $\surd$ &  &  &  &  & $\surd$ & $\surd$ &  &  & NER & BC2GM & 97.27 & 88.38 &   92.61 $\downarrow$\\
 &  & $\surd$ & $\surd$ &  &  & $\surd$ &  &  &  & $\surd$ &  & NER & BC2GM & 95.23 & 92.07 &   93.63 $\downarrow$\\
 &  & $\surd$ & $\surd$ &  &  & $\surd$ &  &  & $\surd$ &  &  & NER & BC2GM & 95.02 & 90.71 &   92.81 $\downarrow$\\
 \hdashline
 &  &  &  & $\surd$ &  &  &  &  &  &  &  & Baseline & BC4CHEMD & 98.62 & 97.20 & 97.91 $\bigstar$ \\
 \hdashline
 &  &  &  & $\surd$ & $\surd$ &  &  &  &  &  &  & NER & BC4CHEMD & 97.36 & 95.64 &   96.49 $\downarrow$\\
 &  &  & $\surd$ & $\surd$ &  &  &  &  &  &  &  & NER & BC4CHEMD & 98.20 & 93.44 &   95.76 $\downarrow$\\
 &  &  &  & $\surd$ &  &  &  &  &  & $\surd$ &  & NER & BC4CHEMD & 97.74 & 91.80 &   94.67 $\downarrow$\\
 & $\surd$ &  &  & $\surd$ &  &  &  &  &  &  &  & NER & BC4CHEMD & 98.02 & 89.97 &   93.82 $\downarrow$\\
 &  &  &  & $\surd$ &  &  &  & $\surd$ &  &  &  & NER & BC4CHEMD & 97.94 & 92.68 &   95.24 $\downarrow$\\
 & $\surd$ &  &  & $\surd$ &  &  &  & $\surd$ &  &  &  & NER & BC4CHEMD & 98.00 & 89.49 &   93.55 $\downarrow$\\
 & $\surd$ &  &  & $\surd$ &  &  &  &  &  & $\surd$ &  & NER & BC4CHEMD & 97.60 & 88.67 &   92.92 $\downarrow$\\
 &  &  &  & $\surd$ &  &  &  & $\surd$ &  & $\surd$ &  & NER & BC4CHEMD & 97.37 & 90.14 &   93.61 $\downarrow$\\
 &  &  & $\surd$ & $\surd$ & $\surd$ &  &  &  &  &  &  & NER & BC4CHEMD & 97.99 & 93.11 &   95.49 $\downarrow$\\
 & $\surd$ &  &  & $\surd$ &  &  &  & $\surd$ &  & $\surd$ &  & NER & BC4CHEMD & 97.49 & 88.01 &   92.51 $\downarrow$\\
$\surd$ &  &  &  & $\surd$ &  & $\surd$ &  &  & $\surd$ &  &  & NER & BC4CHEMD & 98.11 & 87.13 &   92.29 $\downarrow$\\
$\surd$ &  &  &  & $\surd$ &  & $\surd$ &  &  &  & $\surd$ &  & NER & BC4CHEMD & 96.02 & 89.68 &   92.74 $\downarrow$\\
$\surd$ &  &  &  & $\surd$ &  & $\surd$ &  &  &  &  & $\surd$ & NER & BC4CHEMD & 97.94 & 85.96 &   91.56 $\downarrow$\\
$\surd$ &  &  &  & $\surd$ &  &  &  & $\surd$ &  & $\surd$ &  & NER & BC4CHEMD & 97.74 & 88.30 &   92.78 $\downarrow$\\
$\surd$ &  &  &  & $\surd$ &  &  &  & $\surd$ & $\surd$ &  &  & NER & BC4CHEMD & 97.85 & 86.98 &   92.10 $\downarrow$\\
$\surd$ &  &  &  & $\surd$ &  &  &  & $\surd$ &  &  & $\surd$ & NER & BC4CHEMD & 97.96 & 87.50 &   92.43 $\downarrow$\\
$\surd$ &  &  &  & $\surd$ &  &  & $\surd$ &  &  &  & $\surd$ & NER & BC4CHEMD & 96.85 & 86.85 &   91.57 $\downarrow$\\
$\surd$ &  &  &  & $\surd$ &  &  & $\surd$ &  &  & $\surd$ &  & NER & BC4CHEMD & 97.59 & 87.59 &   92.32 $\downarrow$\\
$\surd$ &  &  &  & $\surd$ &  &  & $\surd$ &  & $\surd$ &  &  & NER & BC4CHEMD & 98.12 & 86.31 &   91.84 $\downarrow$\\
 & $\surd$ &  &  & $\surd$ &  & $\surd$ &  &  &  & $\surd$ &  & NER & BC4CHEMD & 96.94 & 88.16 &   92.35 $\downarrow$\\
 & $\surd$ &  &  & $\surd$ &  & $\surd$ &  &  &  &  & $\surd$ & NER & BC4CHEMD & 97.49 & 87.61 &   92.29 $\downarrow$\\
 & $\surd$ &  &  & $\surd$ &  & $\surd$ &  &  & $\surd$ &  &  & NER & BC4CHEMD & 96.90 & 88.23 &   92.36 $\downarrow$\\
 & $\surd$ &  &  & $\surd$ &  &  &  & $\surd$ &  &  & $\surd$ & NER & BC4CHEMD & 97.87 & 86.72 &   91.96 $\downarrow$\\
 &  & $\surd$ &  & $\surd$ &  &  & $\surd$ &  & $\surd$ &  &  & NER & BC4CHEMD & 97.24 & 89.02 &   92.95 $\downarrow$\\
 &  & $\surd$ &  & $\surd$ &  &  & $\surd$ &  &  & $\surd$ &  & NER & BC4CHEMD & 96.51 & 88.46 &   92.31 $\downarrow$\\
 & $\surd$ &  &  & $\surd$ &  &  & $\surd$ &  &  &  & $\surd$ & NER & BC4CHEMD & 98.00 & 86.66 &   91.98 $\downarrow$\\
 &  & $\surd$ &  & $\surd$ &  &  & $\surd$ &  &  &  & $\surd$ & NER & BC4CHEMD & 97.19 & 88.13 &   92.44 $\downarrow$\\
 & $\surd$ &  &  & $\surd$ &  &  & $\surd$ &  & $\surd$ &  &  & NER & BC4CHEMD & 97.86 & 86.43 &   91.79 $\downarrow$\\
 & $\surd$ &  &  & $\surd$ &  &  & $\surd$ &  &  & $\surd$ &  & NER & BC4CHEMD & 97.29 & 87.06 &   91.89 $\downarrow$\\
 &  & $\surd$ &  & $\surd$ &  & $\surd$ &  &  &  & $\surd$ &  & NER & BC4CHEMD & 97.98 & 86.98 &   92.15 $\downarrow$\\
 &  & $\surd$ &  & $\surd$ &  & $\surd$ &  &  & $\surd$ &  &  & NER & BC4CHEMD & 97.39 & 90.75 &   93.96 $\downarrow$\\
 &  & $\surd$ &  & $\surd$ &  & $\surd$ &  &  &  &  & $\surd$ & NER & BC4CHEMD & 97.54 & 88.35 &   92.72 $\downarrow$\\
 &  & $\surd$ &  & $\surd$ &  &  &  & $\surd$ &  & $\surd$ &  & NER & BC4CHEMD & 97.45 & 89.08 &   93.07 $\downarrow$\\
 &  & $\surd$ &  & $\surd$ &  &  &  & $\surd$ & $\surd$ &  &  & NER & BC4CHEMD & 97.59 & 88.92 &   93.06 $\downarrow$\\
 &  & $\surd$ &  & $\surd$ &  &  &  & $\surd$ &  &  & $\surd$ & NER & BC4CHEMD & 98.33 & 87.71 &   92.72 $\downarrow$\\
 \hdashline
 &  &  &  &  & $\surd$ &  &  &  &  &  &  & Baseline & BC5CDR & 97.40 & 95.46 & 96.42 $\bigstar$ \\
 \hdashline
 &  &  &  & $\surd$ & $\surd$ &  &  &  &  &  &  & NER & BC5CDR & 95.98 & 95.06 &   95.52 $\downarrow$\\
 &  &  & $\surd$ &  & $\surd$ &  &  &  &  &  &  & NER & BC5CDR & 96.35 & 95.28 &   95.81 $\downarrow$\\
 &  & $\surd$ &  &  & $\surd$ &  &  &  &  &  &  & NER & BC5CDR & 96.60 & 95.18 &   95.89 $\downarrow$\\
 &  &  &  &  & $\surd$ &  & $\surd$ &  &  &  &  & NER & BC5CDR & 95.80 & 93.96 &   94.87 $\downarrow$\\
 &  &  &  &  & $\surd$ &  &  &  &  &  & $\surd$ & NER & BC5CDR & 96.40 & 94.02 &   95.20 $\downarrow$\\
 &  & $\surd$ &  &  & $\surd$ &  & $\surd$ &  &  &  &  & NER & BC5CDR & 96.10 & 93.54 &   94.80 $\downarrow$\\
 &  & $\surd$ &  &  & $\surd$ &  &  &  &  &  & $\surd$ & NER & BC5CDR & 96.65 & 93.24 &   94.91 $\downarrow$\\
 &  &  &  &  & $\surd$ &  & $\surd$ &  &  &  & $\surd$ & NER & BC5CDR & 96.08 & 91.41 &   93.68 $\downarrow$\\
 &  &  & $\surd$ & $\surd$ & $\surd$ &  &  &  &  &  &  & NER & BC5CDR & 96.74 & 94.21 &   95.46 $\downarrow$\\
 &  & $\surd$ &  &  & $\surd$ &  & $\surd$ &  &  &  & $\surd$ & NER & BC5CDR & 94.68 & 89.67 &   92.11 $\downarrow$\\
$\surd$ &  &  &  &  & $\surd$ & $\surd$ &  &  & $\surd$ &  &  & NER & BC5CDR & 95.11 & 91.20 &   93.11 $\downarrow$\\
$\surd$ &  &  &  &  & $\surd$ & $\surd$ &  &  &  &  & $\surd$ & NER & BC5CDR & 95.46 & 91.06 &   93.21 $\downarrow$\\
$\surd$ &  &  &  &  & $\surd$ & $\surd$ &  &  &  & $\surd$ &  & NER & BC5CDR & 95.37 & 91.38 &   93.33 $\downarrow$\\
$\surd$ &  &  &  &  & $\surd$ &  &  & $\surd$ & $\surd$ &  &  & NER & BC5CDR & 96.63 & 90.16 &   93.28 $\downarrow$\\
$\surd$ &  &  &  &  & $\surd$ &  &  & $\surd$ &  & $\surd$ &  & NER & BC5CDR & 94.83 & 92.23 &   93.51 $\downarrow$\\
$\surd$ &  &  &  &  & $\surd$ &  &  & $\surd$ &  &  & $\surd$ & NER & BC5CDR & 95.87 & 90.12 &   92.90 $\downarrow$\\
$\surd$ &  &  &  &  & $\surd$ &  & $\surd$ &  & $\surd$ &  &  & NER & BC5CDR & 96.00 & 90.96 &   93.41 $\downarrow$\\
$\surd$ &  &  &  &  & $\surd$ &  & $\surd$ &  &  &  & $\surd$ & NER & BC5CDR & 96.44 & 90.55 &   93.40 $\downarrow$\\
$\surd$ &  &  &  &  & $\surd$ &  & $\surd$ &  &  & $\surd$ &  & NER & BC5CDR & 94.25 & 92.09 &   93.15 $\downarrow$\\
 & $\surd$ &  &  &  & $\surd$ & $\surd$ &  &  &  &  & $\surd$ & NER & BC5CDR & 95.66 & 90.59 &   93.06 $\downarrow$\\
 & $\surd$ &  &  &  & $\surd$ & $\surd$ &  &  &  & $\surd$ &  & NER & BC5CDR & 94.48 & 90.76 &   92.58 $\downarrow$\\
 & $\surd$ &  &  &  & $\surd$ & $\surd$ &  &  & $\surd$ &  &  & NER & BC5CDR & 94.09 & 91.93 &   93.00 $\downarrow$\\
 &  & $\surd$ &  &  & $\surd$ &  &  & $\surd$ &  & $\surd$ &  & NER & BC5CDR & 95.74 & 90.61 &   93.10 $\downarrow$\\
 &  & $\surd$ &  &  & $\surd$ &  &  & $\surd$ &  &  & $\surd$ & NER & BC5CDR & 95.30 & 90.90 &   93.05 $\downarrow$\\
 & $\surd$ &  &  &  & $\surd$ &  &  & $\surd$ & $\surd$ &  &  & NER & BC5CDR & 94.81 & 89.67 &   92.17 $\downarrow$\\
 & $\surd$ &  &  &  & $\surd$ &  &  & $\surd$ &  & $\surd$ &  & NER & BC5CDR & 94.68 & 91.31 &   92.97 $\downarrow$\\
 & $\surd$ &  &  &  & $\surd$ &  &  & $\surd$ &  &  & $\surd$ & NER & BC5CDR & 96.04 & 88.27 &   91.99 $\downarrow$\\
 & $\surd$ &  &  &  & $\surd$ &  & $\surd$ &  & $\surd$ &  &  & NER & BC5CDR & 94.67 & 90.13 &   92.34 $\downarrow$\\
 &  & $\surd$ &  &  & $\surd$ &  & $\surd$ &  & $\surd$ &  &  & NER & BC5CDR & 95.42 & 92.04 &   93.70 $\downarrow$\\
 &  & $\surd$ &  &  & $\surd$ &  & $\surd$ &  &  & $\surd$ &  & NER & BC5CDR & 94.01 & 91.52 &   92.75 $\downarrow$\\
 & $\surd$ &  &  &  & $\surd$ &  & $\surd$ &  &  & $\surd$ &  & NER & BC5CDR & 94.66 & 89.73 &   92.13 $\downarrow$\\
 & $\surd$ &  &  &  & $\surd$ &  & $\surd$ &  &  &  & $\surd$ & NER & BC5CDR & 95.65 & 88.83 &   92.11 $\downarrow$\\
 &  & $\surd$ &  &  & $\surd$ & $\surd$ &  &  &  & $\surd$ &  & NER & BC5CDR & 95.71 & 90.97 &   93.28 $\downarrow$\\
 &  & $\surd$ &  &  & $\surd$ & $\surd$ &  &  &  &  & $\surd$ & NER & BC5CDR & 95.77 & 90.18 &   92.89 $\downarrow$\\
 &  & $\surd$ &  &  & $\surd$ & $\surd$ &  &  & $\surd$ &  &  & NER & BC5CDR & 95.62 & 92.18 &   93.87 $\downarrow$\\
 &  & $\surd$ &  &  & $\surd$ &  &  & $\surd$ & $\surd$ &  &  & NER & BC5CDR & 95.76 & 91.35 &   93.50 $\downarrow$\\
\bottomrule
\end{tabular}
}
\label{table:NER_result}
\end{table*}
\end{center}

\begin{center}
\begin{table*}[t]
\centering
\caption{Data preparation for the EE task, including GENIA2011, GENIA2013, and PHEE datasets. $\bigstar$ indicates the best F1 score and best combination in the data preparation phase. Each line represents one experiment and $\surd$ means the dataset is selected for this run.}
\resizebox{\textwidth}{110mm}{
\begin{tabular}{ccccccccccccccccc}
\toprule
\multicolumn{12}{c}{Training sets}\\
\cmidrule(lr){1-12}
\multicolumn{3}{c}{RE task} & \multicolumn{3}{c}{NER task} & \multicolumn{3}{c}{EE task} & \multicolumn{3}{c}{TC task} & & & \multicolumn{3}{c}{Metrics} \\
\cmidrule(lr){1-3} \cmidrule(lr){4-6} \cmidrule(lr){7-9} \cmidrule(lr){10-12} \cmidrule(lr){15-17}
BioRED & DDI & GIT & BC2GM & BC4CHEMD & BC5CDR & GENIA2011 & GENIA2013 & PHEE & ADE & HealthAdvice & PubMed20krct & Task & Test Set & Precision & Recall & F1 Score \\
\midrule
 &  &  &  &  &  & $\surd$ &  &  &  &  &  & Baseline & GENIA2011 & 21.56 & 86.56 & 34.52 \\
 \hdashline
 &  &  &  &  &  & $\surd$ &  & $\surd$ &  &  &  & EE & GENIA2011 & 27.65 & 81.70 &   41.32 $\uparrow$\\
 &  &  &  &  &  & $\surd$ & $\surd$ &  &  &  &  & EE & GENIA2011 & 45.09 & 67.82 &   54.17 $\uparrow$\\
 &  &  &  &  &  & $\surd$ &  &  & $\surd$ &  &  & EE & GENIA2011 & 29.71 & 84.15 &   43.92 $\uparrow$\\
 &  &  & $\surd$ &  &  & $\surd$ &  &  &  &  &  & EE & GENIA2011 & 34.97 & 80.08 &   48.68 $\uparrow$\\
$\surd$ &  &  &  &  &  & $\surd$ &  &  &  &  &  & EE & GENIA2011 & 36.70 & 73.52 &   48.96 $\uparrow$\\
 &  &  &  &  &  & $\surd$ & $\surd$ & $\surd$ &  &  &  & EE & GENIA2011 & 27.45 & 81.06 &   41.01 $\uparrow$\\
$\surd$ &  &  & $\surd$ &  &  & $\surd$ &  &  &  &  &  & EE & GENIA2011 & 28.78 & 82.62 &   42.68 $\uparrow$\\
 &  &  & $\surd$ &  &  & $\surd$ &  &  & $\surd$ &  &  & EE & GENIA2011 & 40.65 & 69.71 &   51.35 $\uparrow$\\
$\surd$ &  &  &  &  &  & $\surd$ &  &  & $\surd$ &  &  & EE & GENIA2011 & 29.95 & 83.68 &   44.11 $\uparrow$\\
$\surd$ &  &  & $\surd$ &  &  & $\surd$ &  &  & $\surd$ &  &  & EE & GENIA2011 & 31.47 & 80.34 &   45.22 $\uparrow$\\
$\surd$ &  &  & $\surd$ &  &  & $\surd$ &  &  &  & $\surd$ &  & EE & GENIA2011 & 29.35 & 80.10 &   42.96 $\uparrow$\\
$\surd$ &  &  & $\surd$ &  &  & $\surd$ &  &  &  &  & $\surd$ & EE & GENIA2011 & 31.83 & 77.86 &   45.19 $\uparrow$\\
$\surd$ &  &  &  & $\surd$ &  & $\surd$ &  &  & $\surd$ &  &  & EE & GENIA2011 & 29.75 & 81.34 &   43.57 $\uparrow$\\
$\surd$ &  &  &  & $\surd$ &  & $\surd$ &  &  &  & $\surd$ &  & EE & GENIA2011 & 28.74 & 82.50 &   42.63 $\uparrow$\\
$\surd$ &  &  &  & $\surd$ &  & $\surd$ &  &  &  &  & $\surd$ & EE & GENIA2011 & 32.13 & 79.31 &   45.73 $\uparrow$\\
$\surd$ &  &  &  &  & $\surd$ & $\surd$ &  &  & $\surd$ &  &  & EE & GENIA2011 & 27.51 & 79.03 &   40.82 $\uparrow$\\
$\surd$ &  &  &  &  & $\surd$ & $\surd$ &  &  &  &  & $\surd$ & EE & GENIA2011 & 27.20 & 83.27 &   41.01 $\uparrow$\\
$\surd$ &  &  &  &  & $\surd$ & $\surd$ &  &  &  & $\surd$ &  & EE & GENIA2011 & 28.94 & 81.48 &   42.71 $\uparrow$\\
 & $\surd$ &  &  &  & $\surd$ & $\surd$ &  &  &  &  & $\surd$ & EE & GENIA2011 & 51.29 & 51.29 &   51.29 $\uparrow$\\
 & $\surd$ &  &  &  & $\surd$ & $\surd$ &  &  &  & $\surd$ &  & EE & GENIA2011 & 42.45 & 64.64 &   51.25 $\uparrow$\\
 & $\surd$ &  & $\surd$ &  &  & $\surd$ &  &  &  &  & $\surd$ & EE & GENIA2011 & 32.99 & 73.61 &   45.56 $\uparrow$\\
 & $\surd$ &  &  & $\surd$ &  & $\surd$ &  &  &  &  & $\surd$ & EE & GENIA2011 & 28.84 & 73.32 &   41.40 $\uparrow$\\
 & $\surd$ &  & $\surd$ &  &  & $\surd$ &  &  &  & $\surd$ &  & EE & GENIA2011 & 34.84 & 57.39 &   43.36 $\uparrow$\\
 & $\surd$ &  &  & $\surd$ &  & $\surd$ &  &  &  & $\surd$ &  & EE & GENIA2011 & 33.04 & 73.99 &   45.68 $\uparrow$\\
 & $\surd$ &  & $\surd$ &  &  & $\surd$ &  &  & $\surd$ &  &  & EE & GENIA2011 & 28.94 & 80.85 &   42.62 $\uparrow$\\
 & $\surd$ &  &  & $\surd$ &  & $\surd$ &  &  & $\surd$ &  &  & EE & GENIA2011 & 28.78 & 69.08 &   40.63 $\uparrow$\\
 & $\surd$ &  &  &  & $\surd$ & $\surd$ &  &  & $\surd$ &  &  & EE & GENIA2011 & 32.81 & 77.67 &   46.13 $\uparrow$\\
 &  & $\surd$ & $\surd$ &  &  & $\surd$ &  &  &  &  & $\surd$ & EE & GENIA2011 & 34.03 & 66.19 &   44.95 $\uparrow$\\
 &  & $\surd$ &  & $\surd$ &  & $\surd$ &  &  &  & $\surd$ &  & EE & GENIA2011 & 35.05 & 76.31 &   48.03 $\uparrow$\\
 &  & $\surd$ &  & $\surd$ &  & $\surd$ &  &  & $\surd$ &  &  & EE & GENIA2011 & 30.67 & 77.54 &   43.95 $\uparrow$\\
 &  & $\surd$ &  & $\surd$ &  & $\surd$ &  &  &  &  & $\surd$ & EE & GENIA2011 & 31.50 & 73.83 &   44.16 $\uparrow$\\
 &  & $\surd$ &  &  & $\surd$ & $\surd$ &  &  &  & $\surd$ &  & EE & GENIA2011 & 38.17 & 69.41 &   49.26 $\uparrow$\\
 &  & $\surd$ &  &  & $\surd$ & $\surd$ &  &  & $\surd$ &  &  & EE & GENIA2011 & 31.54 & 81.54 &   45.48 $\uparrow$\\
 &  & $\surd$ &  &  & $\surd$ & $\surd$ &  &  &  &  & $\surd$ & EE & GENIA2011 & 53.91 & 60.65 &   57.08 $\bigstar$ \\
 &  & $\surd$ & $\surd$ &  &  & $\surd$ &  &  &  & $\surd$ &  & EE & GENIA2011 & 33.08 & 78.30 &   46.51 $\uparrow$\\
 &  & $\surd$ & $\surd$ &  &  & $\surd$ &  &  & $\surd$ &  &  & EE & GENIA2011 & 32.51 & 77.04 &   45.73 $\uparrow$\\
 \hdashline
 &  &  &  &  &  &  & $\surd$ &  &  &  &  & Baseline & GENIA2013 & 14.42 & 37.54 & 20.83 \\
 \hdashline
 &  &  &  &  &  & $\surd$ & $\surd$ &  &  &  &  & EE & GENIA2013 & 13.61 & 35.49 &   19.68 $\downarrow$\\
 &  &  &  &  &  &  & $\surd$ & $\surd$ &  &  &  & EE & GENIA2013 & 19.62 & 48.81 &   27.98 $\uparrow$\\
 &  &  &  &  & $\surd$ &  & $\surd$ &  &  &  &  & EE & GENIA2013 & 9.31 & 27.99 &   13.97 $\downarrow$\\
 &  & $\surd$ &  &  &  &  & $\surd$ &  &  &  &  & EE & GENIA2013 & 16.89 & 43.34 &   24.31 $\uparrow$\\
 &  &  &  &  &  &  & $\surd$ &  &  &  & $\surd$ & EE & GENIA2013 & 16.49 & 42.66 &   23.79 $\uparrow$\\
 &  &  &  &  &  & $\surd$ & $\surd$ & $\surd$ &  &  &  & EE & GENIA2013 & 15.37 & 45.73 &   23.00 $\uparrow$\\
 &  & $\surd$ &  &  & $\surd$ &  & $\surd$ &  &  &  &  & EE & GENIA2013 & 13.08 & 39.25 &   19.62 $\downarrow$$\uparrow$\\
 &  & $\surd$ &  &  &  &  & $\surd$ &  &  &  & $\surd$ & EE & GENIA2013 & 28.75 & 61.43 &   39.17 $\bigstar$ \\
 &  &  &  &  & $\surd$ &  & $\surd$ &  &  &  & $\surd$ & EE & GENIA2013 & 7.20 & 23.55 &   11.03 $\downarrow$\\
 &  & $\surd$ &  &  & $\surd$ &  & $\surd$ &  &  &  & $\surd$ & EE & GENIA2013 & 24.76 & 60.41 &   35.12 $\uparrow$\\
$\surd$ &  &  & $\surd$ &  &  &  & $\surd$ &  & $\surd$ &  &  & EE & GENIA2013 & 16.75 & 45.39 &   24.47 $\uparrow$\\
$\surd$ &  &  & $\surd$ &  &  &  & $\surd$ &  &  & $\surd$ &  & EE & GENIA2013 & 15.59 & 47.10 &   23.43 $\uparrow$\\
$\surd$ &  &  &  & $\surd$ &  &  & $\surd$ &  &  & $\surd$ &  & EE & GENIA2013 & 20.14 & 69.28 &   31.21 $\uparrow$\\
$\surd$ &  &  &  & $\surd$ &  &  & $\surd$ &  & $\surd$ &  &  & EE & GENIA2013 & 9.22 & 33.11 &   14.42 $\downarrow$\\
$\surd$ &  &  &  & $\surd$ &  &  & $\surd$ &  &  &  & $\surd$ & EE & GENIA2013 & 22.08 & 53.58 &   31.27 $\uparrow$\\
$\surd$ &  &  & $\surd$ &  &  &  & $\surd$ &  &  &  & $\surd$ & EE & GENIA2013 & 20.69 & 59.39 &   30.69 $\uparrow$\\
$\surd$ &  &  &  &  & $\surd$ &  & $\surd$ &  & $\surd$ &  &  & EE & GENIA2013 & 9.46 & 31.40 &   14.53 $\downarrow$\\
$\surd$ &  &  &  &  & $\surd$ &  & $\surd$ &  &  & $\surd$ &  & EE & GENIA2013 & 19.35 & 65.19 &   29.84 $\uparrow$\\
$\surd$ &  &  &  &  & $\surd$ &  & $\surd$ &  &  &  & $\surd$ & EE & GENIA2013 & 17.22 & 47.44 &   25.27 $\uparrow$\\
 &  & $\surd$ & $\surd$ &  &  &  & $\surd$ &  &  & $\surd$ &  & EE & GENIA2013 & 24.10 & 59.73 &   34.35 $\uparrow$\\
 &  & $\surd$ & $\surd$ &  &  &  & $\surd$ &  & $\surd$ &  &  & EE & GENIA2013 & 12.70 & 35.84 &   18.75 $\downarrow$\\
 &  & $\surd$ & $\surd$ &  &  &  & $\surd$ &  &  &  & $\surd$ & EE & GENIA2013 & 15.05 & 39.46 &   21.78 $\uparrow$\\
 &  & $\surd$ &  & $\surd$ &  &  & $\surd$ &  & $\surd$ &  &  & EE & GENIA2013 & 8.60 & 30.61 &   13.43 $\downarrow$\\
 & $\surd$ &  & $\surd$ &  &  &  & $\surd$ &  &  &  & $\surd$ & EE & GENIA2013 & 17.49 & 47.10 &   25.51 $\uparrow$\\
 & $\surd$ &  & $\surd$ &  &  &  & $\surd$ &  & $\surd$ &  &  & EE & GENIA2013 & 16.48 & 49.83 &   24.77 $\uparrow$\\
 &  & $\surd$ &  & $\surd$ &  &  & $\surd$ &  &  & $\surd$ &  & EE & GENIA2013 & 11.14 & 34.81 &   16.87 $\downarrow$\\
 &  & $\surd$ &  & $\surd$ &  &  & $\surd$ &  &  &  & $\surd$ & EE & GENIA2013 & 26.68 & 70.31 &   38.69 $\uparrow$\\
 & $\surd$ &  & $\surd$ &  &  &  & $\surd$ &  &  & $\surd$ &  & EE & GENIA2013 & 21.57 & 64.85 &   32.37 $\uparrow$\\
 & $\surd$ &  &  & $\surd$ &  &  & $\surd$ &  &  &  & $\surd$ & EE & GENIA2013 & 22.03 & 59.39 &   32.13 $\uparrow$\\
 & $\surd$ &  &  &  & $\surd$ &  & $\surd$ &  & $\surd$ &  &  & EE & GENIA2013 & 18.03 & 57.00 &   27.40 $\uparrow$\\
 & $\surd$ &  &  & $\surd$ &  &  & $\surd$ &  & $\surd$ &  &  & EE & GENIA2013 & 12.66 & 40.27 &   19.27 $\uparrow$\\
 & $\surd$ &  &  & $\surd$ &  &  & $\surd$ &  &  & $\surd$ &  & EE & GENIA2013 & 19.19 & 53.24 &   28.21 $\uparrow$\\
 &  & $\surd$ &  &  & $\surd$ &  & $\surd$ &  & $\surd$ &  &  & EE & GENIA2013 & 17.22 & 44.37 &   24.81 $\uparrow$\\
 &  & $\surd$ &  &  & $\surd$ &  & $\surd$ &  &  & $\surd$ &  & EE & GENIA2013 & 21.78 & 52.56 &   30.80 $\uparrow$\\
 & $\surd$ &  &  &  & $\surd$ &  & $\surd$ &  &  & $\surd$ &  & EE & GENIA2013 & 17.19 & 64.63 &   27.16 $\uparrow$\\
 & $\surd$ &  &  &  & $\surd$ &  & $\surd$ &  &  &  & $\surd$ & EE & GENIA2013 & 15.52 & 42.32 &   22.71 $\uparrow$\\
 \hdashline
 &  &  &  &  &  &  &  & $\surd$ &  &  &  & Baseline & PHEE & 48.51 & 93.75 & 63.94 $\bigstar$ \\
 \hdashline
 &  &  &  &  &  &  & $\surd$ & $\surd$ &  &  &  & EE & PHEE & 55.56 & 93.15 &   69.60 $\uparrow$\\
 &  &  &  &  &  & $\surd$ &  & $\surd$ &  &  &  & EE & PHEE & 47.93 & 93.55 &   63.39 $\downarrow$\\
 &  &  &  &  &  &  &  & $\surd$ &  & $\surd$ &  & EE & PHEE & 49.73 & 92.34 &   64.64 $\downarrow$\\
 &  &  &  & $\surd$ &  &  &  & $\surd$ &  &  &  & EE & PHEE & 45.35 & 84.48 &   59.01 $\downarrow$\\
 & $\surd$ &  &  &  &  &  &  & $\surd$ &  &  &  & EE & PHEE & 76.19 & 89.01 &   82.10 $\uparrow$\\
 &  &  &  &  &  & $\surd$ & $\surd$ & $\surd$ &  &  &  & EE & PHEE & 47.55 & 92.84 &   62.89 $\downarrow$\\
 & $\surd$ &  &  & $\surd$ &  &  &  & $\surd$ &  &  &  & EE & PHEE & 46.41 & 90.02 &   61.25 $\downarrow$\\
 & $\surd$ &  &  &  &  &  &  & $\surd$ &  & $\surd$ &  & EE & PHEE & 47.36 & 92.24 &   62.59 $\downarrow$\\
 &  &  &  & $\surd$ &  &  &  & $\surd$ &  & $\surd$ &  & EE & PHEE & 47.39 & 92.34 &   62.63 $\downarrow$\\
 & $\surd$ &  &  & $\surd$ &  &  &  & $\surd$ &  & $\surd$ &  & EE & PHEE & 69.35 & 88.51 &   77.77 $\uparrow$\\
$\surd$ &  &  & $\surd$ &  &  &  &  & $\surd$ & $\surd$ &  &  & EE & PHEE & 47.67 & 92.84 &   63.00 $\downarrow$\\
$\surd$ &  &  & $\surd$ &  &  &  &  & $\surd$ &  & $\surd$ &  & EE & PHEE & 47.62 & 92.94 &   62.98 $\downarrow$\\
$\surd$ &  &  &  & $\surd$ &  &  &  & $\surd$ & $\surd$ &  &  & EE & PHEE & 47.37 & 92.44 &   62.64 $\downarrow$\\
$\surd$ &  &  &  & $\surd$ &  &  &  & $\surd$ &  & $\surd$ &  & EE & PHEE & 47.49 & 92.64 &   62.79 $\downarrow$\\
$\surd$ &  &  &  & $\surd$ &  &  &  & $\surd$ &  &  & $\surd$ & EE & PHEE & 49.86 & 92.44 &   64.78 $\downarrow$\\
$\surd$ &  &  &  &  & $\surd$ &  &  & $\surd$ & $\surd$ &  &  & EE & PHEE & 47.10 & 91.53 &   62.19 $\downarrow$\\
$\surd$ &  &  & $\surd$ &  &  &  &  & $\surd$ &  &  & $\surd$ & EE & PHEE & 47.42 & 92.54 &   62.70 $\downarrow$\\
$\surd$ &  &  &  &  & $\surd$ &  &  & $\surd$ &  & $\surd$ &  & EE & PHEE & 49.35 & 91.43 &   64.10 $\downarrow$\\
$\surd$ &  &  &  &  & $\surd$ &  &  & $\surd$ &  &  & $\surd$ & EE & PHEE & 47.78 & 93.15 &   63.16 $\downarrow$\\
 & $\surd$ &  & $\surd$ &  &  &  &  & $\surd$ & $\surd$ &  &  & EE & PHEE & 46.53 & 90.62 &   61.49 $\downarrow$\\
 &  & $\surd$ &  &  & $\surd$ &  &  & $\surd$ &  & $\surd$ &  & EE & PHEE & 51.09 & 89.82 &   65.13 $\downarrow$\\
 &  & $\surd$ &  &  & $\surd$ &  &  & $\surd$ &  &  & $\surd$ & EE & PHEE & 47.83 & 79.94 &   59.85 $\downarrow$\\
 & $\surd$ &  & $\surd$ &  &  &  &  & $\surd$ &  & $\surd$ &  & EE & PHEE & 46.07 & 89.31 &   60.79 $\downarrow$\\
 & $\surd$ &  & $\surd$ &  &  &  &  & $\surd$ &  &  & $\surd$ & EE & PHEE & 18.18 & 21.77 &   19.82 $\downarrow$\\
 & $\surd$ &  &  &  & $\surd$ &  &  & $\surd$ & $\surd$ &  &  & EE & PHEE & 53.36 & 67.24 &   59.50 $\downarrow$\\
 & $\surd$ &  &  &  & $\surd$ &  &  & $\surd$ &  & $\surd$ &  & EE & PHEE & 47.67 & 92.94 &   63.02 $\downarrow$\\
 & $\surd$ &  &  &  & $\surd$ &  &  & $\surd$ &  &  & $\surd$ & EE & PHEE & 49.23 & 90.83 &   63.86 $\downarrow$\\
 & $\surd$ &  &  & $\surd$ &  &  &  & $\surd$ &  &  & $\surd$ & EE & PHEE & 48.48 & 93.04 &   63.74 $\downarrow$\\
 &  & $\surd$ &  & $\surd$ &  &  &  & $\surd$ & $\surd$ &  &  & EE & PHEE & 46.95 & 90.02 &   61.71 $\downarrow$\\
 &  & $\surd$ & $\surd$ &  &  &  &  & $\surd$ &  & $\surd$ &  & EE & PHEE & 48.53 & 91.63 &   63.46 $\downarrow$\\
 &  & $\surd$ &  & $\surd$ &  &  &  & $\surd$ &  & $\surd$ &  & EE & PHEE & 94.23 & 92.14 &   93.17 $\uparrow$\\
 &  & $\surd$ & $\surd$ &  &  &  &  & $\surd$ & $\surd$ &  &  & EE & PHEE & 48.32 & 92.84 &   63.56 $\downarrow$\\
 &  & $\surd$ &  & $\surd$ &  &  &  & $\surd$ &  &  & $\surd$ & EE & PHEE & 47.64 & 92.64 &   62.92 $\downarrow$\\
 &  & $\surd$ & $\surd$ &  &  &  &  & $\surd$ &  &  & $\surd$ & EE & PHEE & 53.27 & 95.97 &   68.51 $\downarrow$\\
 &  & $\surd$ &  &  & $\surd$ &  &  & $\surd$ & $\surd$ &  &  & EE & PHEE & 47.72 & 93.04 &   63.09 $\downarrow$\\
\bottomrule
\end{tabular}
}
\label{table:EE_result}
\end{table*}
\end{center}

\begin{center}
\begin{table*}[t]
\centering
\caption{Data preparation for the TC task, including ADE, HealthAdvice, and PubMed20krct datasets. $\bigstar$ indicates the best F1 score and best combination in the data preparation phase. Each line represents one experiment and $\surd$ means the dataset is selected for this run.}
\resizebox{\textwidth}{110mm}{
\begin{tabular}{ccccccccccccccccc}
\toprule
\multicolumn{12}{c}{Training sets}\\
\cmidrule(lr){1-12}
\multicolumn{3}{c}{RE task} & \multicolumn{3}{c}{NER task} & \multicolumn{3}{c}{EE task} & \multicolumn{3}{c}{TC task} & & & \multicolumn{3}{c}{Metrics} \\
\cmidrule(lr){1-3} \cmidrule(lr){4-6} \cmidrule(lr){7-9} \cmidrule(lr){10-12} \cmidrule(lr){15-17}
BioRED & DDI & GIT & BC2GM & BC4CHEMD & BC5CDR & GENIA2011 & GENIA2013 & PHEE & ADE & HealthAdvice & PubMed20krct & Task & Test Set & Precision & Recall & F1 Score \\
\midrule
  &  &  &  &  &  &  &  &  & $\surd$ &  &  & TC & ADE & 93.20 & 93.20 & 93.20 $\bigstar$ \\
  \hdashline
 &  &  &  &  &  &  &  &  & $\surd$ & $\surd$ &  & TC & ADE & 80.30 & 80.30 &   80.30 $\downarrow$\\
 &  &  &  &  &  &  &  &  & $\surd$ &  & $\surd$ & TC & ADE & 59.40 & 59.40 &   59.40 $\downarrow$\\
 &  &  & $\surd$ &  &  &  &  &  & $\surd$ &  &  & TC & ADE & 90.10 & 90.10 &   90.10 $\downarrow$\\
 &  &  &  &  &  & $\surd$ &  &  & $\surd$ &  &  & TC & ADE & 60.28 & 89.40 &   72.01 $\downarrow$\\
$\surd$ &  &  &  &  &  &  &  &  & $\surd$ &  &  & TC & ADE & 81.31 & 88.30 &   84.66 $\downarrow$\\
 &  &  &  &  &  &  &  &  & $\surd$ & $\surd$ & $\surd$ & TC & ADE & 86.90 & 86.90 &   86.90 $\downarrow$\\
$\surd$ &  &  &  &  &  & $\surd$ &  &  & $\surd$ &  &  & TC & ADE & 72.40 & 89.70 &   80.13 $\downarrow$\\
 &  &  & $\surd$ &  &  & $\surd$ &  &  & $\surd$ &  &  & TC & ADE & 86.60 & 89.20 &   87.88 $\downarrow$\\
$\surd$ &  &  & $\surd$ &  &  &  &  &  & $\surd$ &  &  & TC & ADE & 46.89 & 87.30 &   61.01 $\downarrow$\\
$\surd$ &  &  & $\surd$ &  &  & $\surd$ &  &  & $\surd$ &  &  & TC & ADE & 36.46 & 84.00 &   50.85 $\downarrow$\\
 &  & $\surd$ &  &  & $\surd$ &  & $\surd$ &  & $\surd$ &  &  & TC & ADE & 85.00 & 85.00 &   85.00 $\downarrow$\\
 &  & $\surd$ &  & $\surd$ &  &  & $\surd$ &  & $\surd$ &  &  & TC & ADE & 89.20 & 89.20 &   89.20 $\downarrow$\\
 &  & $\surd$ &  & $\surd$ &  & $\surd$ &  &  & $\surd$ &  &  & TC & ADE & 56.42 & 80.80 &   66.45 $\downarrow$\\
 &  & $\surd$ &  & $\surd$ &  &  &  & $\surd$ & $\surd$ &  &  & TC & ADE & 68.17 & 70.90 &   69.51 $\downarrow$\\
 &  & $\surd$ & $\surd$ &  &  &  & $\surd$ &  & $\surd$ &  &  & TC & ADE & 87.30 & 87.30 &   87.30 $\downarrow$\\
$\surd$ &  &  &  &  & $\surd$ &  & $\surd$ &  & $\surd$ &  &  & TC & ADE & 83.00 & 83.00 &   83.00 $\downarrow$\\
 &  & $\surd$ & $\surd$ &  &  &  &  & $\surd$ & $\surd$ &  &  & TC & ADE & 83.47 & 83.80 &   83.63 $\downarrow$\\
 & $\surd$ &  & $\surd$ &  &  &  & $\surd$ &  & $\surd$ &  &  & TC & ADE & 82.30 & 82.30 &   82.30 $\downarrow$\\
$\surd$ &  &  &  &  & $\surd$ & $\surd$ &  &  & $\surd$ &  &  & TC & ADE & 29.17 & 33.20 &   31.06 $\downarrow$\\
 &  & $\surd$ & $\surd$ &  &  & $\surd$ &  &  & $\surd$ &  &  & TC & ADE & 44.69 & 90.80 &   59.89 $\downarrow$\\
 & $\surd$ &  &  & $\surd$ &  &  & $\surd$ &  & $\surd$ &  &  & TC & ADE & 79.80 & 79.80 &   79.80 $\downarrow$\\
$\surd$ &  &  &  &  & $\surd$ &  &  & $\surd$ & $\surd$ &  &  & TC & ADE & 30.24 & 33.90 &   31.97 $\downarrow$\\
 & $\surd$ &  &  & $\surd$ &  & $\surd$ &  &  & $\surd$ &  &  & TC & ADE & 23.99 & 32.50 &   27.60 $\downarrow$\\
 & $\surd$ &  &  &  & $\surd$ &  &  & $\surd$ & $\surd$ &  &  & TC & ADE & 87.60 & 87.60 &   87.60 $\downarrow$\\
 & $\surd$ &  &  &  & $\surd$ &  & $\surd$ &  & $\surd$ &  &  & TC & ADE & 84.50 & 84.50 &   84.50 $\downarrow$\\
 &  & $\surd$ &  &  & $\surd$ & $\surd$ &  &  & $\surd$ &  &  & TC & ADE & 55.58 & 89.60 &   68.61 $\downarrow$\\
 &  & $\surd$ &  &  & $\surd$ &  &  & $\surd$ & $\surd$ &  &  & TC & ADE & 89.81 & 89.90 &   89.86 $\downarrow$\\
$\surd$ &  &  &  & $\surd$ &  & $\surd$ &  &  & $\surd$ &  &  & TC & ADE & 48.14 & 53.00 &   50.45 $\downarrow$\\
$\surd$ &  &  & $\surd$ &  &  &  & $\surd$ &  & $\surd$ &  &  & TC & ADE & 80.71 & 82.00 &   81.35 $\downarrow$\\
$\surd$ &  &  & $\surd$ &  &  &  &  & $\surd$ & $\surd$ &  &  & TC & ADE & 67.46 & 68.00 &   67.73 $\downarrow$\\
$\surd$ &  &  &  & $\surd$ &  &  &  & $\surd$ & $\surd$ &  &  & TC & ADE & 40.39 & 49.20 &   44.36 $\downarrow$\\
$\surd$ &  &  &  & $\surd$ &  &  & $\surd$ &  & $\surd$ &  &  & TC & ADE & 80.40 & 80.40 &   80.40 $\downarrow$\\
 & $\surd$ &  &  &  & $\surd$ & $\surd$ &  &  & $\surd$ &  &  & TC & ADE & 86.80 & 86.80 &   86.80 $\downarrow$\\
 & $\surd$ &  & $\surd$ &  &  & $\surd$ &  &  & $\surd$ &  &  & TC & ADE & 59.61 & 78.80 &   67.87 $\downarrow$\\
 & $\surd$ &  & $\surd$ &  &  &  &  & $\surd$ & $\surd$ &  &  & TC & ADE & 70.58 & 72.70 &   71.63 $\downarrow$\\
 \hdashline
 &  &  &  &  &  &  &  &  &  & $\surd$ &  & TC & HealthAdvice & 91.00 & 91.00 & 91.00 $\bigstar$ \\
 \hdashline
 &  &  &  &  &  &  &  &  & $\surd$ & $\surd$ &  & TC & HealthAdvice & 89.70 & 89.70 &   89.70 $\downarrow$\\
 &  &  &  &  &  &  &  &  &  & $\surd$ & $\surd$ & TC & HealthAdvice & 90.10 & 90.10 &   90.10 $\downarrow$\\
 & $\surd$ &  &  &  &  &  &  &  &  & $\surd$ &  & TC & HealthAdvice & 87.30 & 87.30 &   87.30 $\downarrow$\\
 &  &  &  &  &  &  &  & $\surd$ &  & $\surd$ &  & TC & HealthAdvice & 83.93 & 86.70 &   85.29 $\downarrow$\\
 &  &  &  & $\surd$ &  &  &  &  &  & $\surd$ &  & TC & HealthAdvice & 85.70 & 85.70 &   85.70 $\downarrow$\\
 &  &  &  &  &  &  &  &  & $\surd$ & $\surd$ & $\surd$ & TC & HealthAdvice & 87.20 & 87.20 &   87.20 $\downarrow$\\
 & $\surd$ &  &  &  &  &  &  & $\surd$ &  & $\surd$ &  & TC & HealthAdvice & 79.74 & 80.30 &   80.02 $\downarrow$\\
 & $\surd$ &  &  & $\surd$ &  &  &  &  &  & $\surd$ &  & TC & HealthAdvice & 84.80 & 84.80 &   84.80 $\downarrow$\\
 &  &  &  & $\surd$ &  &  &  & $\surd$ &  & $\surd$ &  & TC & HealthAdvice & 62.05 & 83.40 &   71.16 $\downarrow$\\
 & $\surd$ &  &  & $\surd$ &  &  &  & $\surd$ &  & $\surd$ &  & TC & HealthAdvice & 77.60 & 77.60 &   77.60 $\downarrow$\\
 &  & $\surd$ &  &  & $\surd$ &  & $\surd$ &  &  & $\surd$ &  & TC & HealthAdvice & 85.20 & 85.20 &   85.20 $\downarrow$\\
 &  & $\surd$ &  & $\surd$ &  &  & $\surd$ &  &  & $\surd$ &  & TC & HealthAdvice & 85.20 & 85.20 &   85.20 $\downarrow$\\
 &  & $\surd$ & $\surd$ &  &  & $\surd$ &  &  &  & $\surd$ &  & TC & HealthAdvice & 78.51 & 81.10 &   79.78 $\downarrow$\\
 &  & $\surd$ &  & $\surd$ &  &  &  & $\surd$ &  & $\surd$ &  & TC & HealthAdvice & 83.50 & 83.50 &   83.50 $\downarrow$\\
 &  & $\surd$ &  & $\surd$ &  & $\surd$ &  &  &  & $\surd$ &  & TC & HealthAdvice & 53.56 & 81.30 &   64.58 $\downarrow$\\
 &  & $\surd$ & $\surd$ &  &  &  & $\surd$ &  &  & $\surd$ &  & TC & HealthAdvice & 86.50 & 86.50 &   86.50 $\downarrow$\\
 & $\surd$ &  &  & $\surd$ &  &  & $\surd$ &  &  & $\surd$ &  & TC & HealthAdvice & 55.81 & 80.20 &   65.82 $\downarrow$\\
 &  & $\surd$ & $\surd$ &  &  &  &  & $\surd$ &  & $\surd$ &  & TC & HealthAdvice & 80.06 & 84.70 &   82.31 $\downarrow$\\
$\surd$ &  &  &  &  & $\surd$ & $\surd$ &  &  &  & $\surd$ &  & TC & HealthAdvice & 41.20 & 92.50 &   57.01 $\downarrow$\\
 &  & $\surd$ &  &  & $\surd$ &  &  & $\surd$ &  & $\surd$ &  & TC & HealthAdvice & 84.80 & 84.80 &   84.80 $\downarrow$\\
 &  & $\surd$ &  &  & $\surd$ & $\surd$ &  &  &  & $\surd$ &  & TC & HealthAdvice & 82.00 & 87.90 &   84.85 $\downarrow$\\
 & $\surd$ &  & $\surd$ &  &  &  & $\surd$ &  &  & $\surd$ &  & TC & HealthAdvice & 84.33 & 85.60 &   84.96 $\downarrow$\\
$\surd$ &  &  &  &  & $\surd$ &  &  & $\surd$ &  & $\surd$ &  & TC & HealthAdvice & 86.90 & 86.90 &   86.90 $\downarrow$\\
 & $\surd$ &  & $\surd$ &  &  & $\surd$ &  &  &  & $\surd$ &  & TC & HealthAdvice & 59.73 & 78.60 &   67.88 $\downarrow$\\
 & $\surd$ &  &  &  & $\surd$ &  &  & $\surd$ &  & $\surd$ &  & TC & HealthAdvice & 77.30 & 77.30 &   77.30 $\downarrow$\\
 & $\surd$ &  & $\surd$ &  &  &  &  & $\surd$ &  & $\surd$ &  & TC & HealthAdvice & 84.40 & 84.40 &   84.40 $\downarrow$\\
 & $\surd$ &  &  &  & $\surd$ &  & $\surd$ &  &  & $\surd$ &  & TC & HealthAdvice & 81.50 & 81.50 &   81.50 $\downarrow$\\
$\surd$ &  &  &  &  & $\surd$ &  & $\surd$ &  &  & $\surd$ &  & TC & HealthAdvice & 57.08 & 83.40 &   67.78 $\downarrow$\\
$\surd$ &  &  &  & $\surd$ &  & $\surd$ &  &  &  & $\surd$ &  & TC & HealthAdvice & 69.02 & 80.20 &   74.19 $\downarrow$\\
 & $\surd$ &  &  &  & $\surd$ & $\surd$ &  &  &  & $\surd$ &  & TC & HealthAdvice & 82.92 & 83.00 &   82.96 $\downarrow$\\
$\surd$ &  &  & $\surd$ &  &  & $\surd$ &  &  &  & $\surd$ &  & TC & HealthAdvice & 83.01 & 85.00 &   83.99 $\downarrow$\\
$\surd$ &  &  & $\surd$ &  &  &  & $\surd$ &  &  & $\surd$ &  & TC & HealthAdvice & 84.32 & 84.40 &   84.36 $\downarrow$\\
$\surd$ &  &  &  & $\surd$ &  &  &  & $\surd$ &  & $\surd$ &  & TC & HealthAdvice & 68.04 & 72.80 &   70.34 $\downarrow$\\
$\surd$ &  &  & $\surd$ &  &  &  &  & $\surd$ &  & $\surd$ &  & TC & HealthAdvice & 78.00 & 78.00 &   78.00 $\downarrow$\\
$\surd$ &  &  &  & $\surd$ &  &  & $\surd$ &  &  & $\surd$ &  & TC & HealthAdvice & 58.51 & 85.60 &   69.51 $\downarrow$\\
 & $\surd$ &  &  & $\surd$ &  & $\surd$ &  &  &  & $\surd$ &  & TC & HealthAdvice & 83.17 & 84.50 &   83.83 $\downarrow$\\
 \hdashline
 &  &  &  &  &  &  &  &  &  &  & $\surd$ & TC & PubMed20krct & 87.30 & 87.30 & 87.30 $\bigstar$
\\
 \hdashline
 &  &  &  &  &  &  &  &  & $\surd$ &  & $\surd$ & TC & PubMed20krct & 84.60 & 84.60 &   84.60 $\downarrow$\\
 &  &  &  &  &  &  &  &  &  & $\surd$ & $\surd$ & TC & PubMed20krct & 83.10 & 83.10 &   83.10 $\downarrow$\\
 &  & $\surd$ &  &  &  &  &  &  &  &  & $\surd$ & TC & PubMed20krct & 84.20 & 84.20 &   84.20 $\downarrow$\\
 &  &  &  &  & $\surd$ &  &  &  &  &  & $\surd$ & TC & PubMed20krct & 85.10 & 85.10 &   85.10 $\downarrow$\\
 &  &  &  &  &  &  & $\surd$ &  &  &  & $\surd$ & TC & PubMed20krct & 84.36 & 84.70 &   84.53 $\downarrow$\\
 &  &  &  &  &  &  &  &  & $\surd$ & $\surd$ & $\surd$ & TC & PubMed20krct & 85.10 & 85.10 &   85.10 $\downarrow$\\
 &  & $\surd$ &  &  &  &  & $\surd$ &  &  &  & $\surd$ & TC & PubMed20krct & 82.28 & 83.10 &   82.69 $\downarrow$\\
 &  & $\surd$ &  &  & $\surd$ &  &  &  &  &  & $\surd$ & TC & PubMed20krct & 86.30 & 86.30 &   86.30 $\downarrow$\\
 &  &  &  &  & $\surd$ &  & $\surd$ &  &  &  & $\surd$ & TC & PubMed20krct & 82.64 & 83.30 &   82.97 $\downarrow$\\
 &  & $\surd$ &  &  & $\surd$ &  & $\surd$ &  &  &  & $\surd$ & TC & PubMed20krct & 81.40 & 81.40 &   81.40 $\downarrow$\\
 &  & $\surd$ & $\surd$ &  &  &  & $\surd$ &  &  &  & $\surd$ & TC & PubMed20krct & 82.90 & 82.90 &   82.90 $\downarrow$\\
 &  & $\surd$ &  & $\surd$ &  &  & $\surd$ &  &  &  & $\surd$ & TC & PubMed20krct & 84.10 & 84.10 &   84.10 $\downarrow$\\
 &  & $\surd$ &  &  & $\surd$ & $\surd$ &  &  &  &  & $\surd$ & TC & PubMed20krct & 84.30 & 84.30 &   84.30 $\downarrow$\\
 &  & $\surd$ & $\surd$ &  &  & $\surd$ &  &  &  &  & $\surd$ & TC & PubMed20krct & 82.02 & 83.00 &   82.50 $\downarrow$\\
 & $\surd$ &  &  & $\surd$ &  &  & $\surd$ &  &  &  & $\surd$ & TC & PubMed20krct & 56.46 & 59.90 &   58.13 $\downarrow$\\
 &  & $\surd$ &  & $\surd$ &  & $\surd$ &  &  &  &  & $\surd$ & TC & PubMed20krct & 75.84 & 81.60 &   78.61 $\downarrow$\\
 &  & $\surd$ & $\surd$ &  &  &  &  & $\surd$ &  &  & $\surd$ & TC & PubMed20krct & 84.40 & 84.40 &   84.40 $\downarrow$\\
 &  & $\surd$ &  & $\surd$ &  &  &  & $\surd$ &  &  & $\surd$ & TC & PubMed20krct & 69.37 & 71.10 &   70.22 $\downarrow$\\
 &  & $\surd$ &  &  & $\surd$ &  &  & $\surd$ &  &  & $\surd$ & TC & PubMed20krct & 81.30 & 81.30 &   81.30 $\downarrow$\\
$\surd$ &  &  &  &  & $\surd$ & $\surd$ &  &  &  &  & $\surd$ & TC & PubMed20krct & 48.16 & 83.60 &   61.11 $\downarrow$\\
$\surd$ &  &  &  &  & $\surd$ &  &  & $\surd$ &  &  & $\surd$ & TC & PubMed20krct & 48.26 & 72.00 &   57.78 $\downarrow$\\
 & $\surd$ &  &  &  & $\surd$ &  &  & $\surd$ &  &  & $\surd$ & TC & PubMed20krct & 58.81 & 61.40 &   60.08 $\downarrow$\\
 & $\surd$ &  &  &  & $\surd$ &  & $\surd$ &  &  &  & $\surd$ & TC & PubMed20krct & 69.57 & 72.00 &   70.76 $\downarrow$\\
 & $\surd$ &  & $\surd$ &  &  &  &  & $\surd$ &  &  & $\surd$ & TC & PubMed20krct & 82.90 & 82.90 &   82.90 $\downarrow$\\
 & $\surd$ &  &  & $\surd$ &  &  &  & $\surd$ &  &  & $\surd$ & TC & PubMed20krct & 59.93 & 68.80 &   64.06 $\downarrow$\\
 & $\surd$ &  & $\surd$ &  &  &  & $\surd$ &  &  &  & $\surd$ & TC & PubMed20krct & 66.98 & 78.90 &   72.45 $\downarrow$\\
$\surd$ &  &  &  &  & $\surd$ &  & $\surd$ &  &  &  & $\surd$ & TC & PubMed20krct & 82.30 & 82.30 &     82.30 $\downarrow$\\
 & $\surd$ &  &  &  & $\surd$ & $\surd$ &  &  &  &  & $\surd$ & TC & PubMed20krct & 75.75 & 77.80 &   76.76 $\downarrow$\\
$\surd$ &  &  & $\surd$ &  &  & $\surd$ &  &  &  &  & $\surd$ & TC & PubMed20krct & 49.03 & 78.30 &   60.30 $\downarrow$\\
$\surd$ &  &  &  & $\surd$ &  & $\surd$ &  &  &  &  & $\surd$ & TC & PubMed20krct & 34.66 & 80.40 &   48.43 $\downarrow$\\
$\surd$ &  &  & $\surd$ &  &  &  & $\surd$ &  &  &  & $\surd$ & TC & PubMed20krct & 71.18 & 72.60 &   71.88 $\downarrow$\\
$\surd$ &  &  &  & $\surd$ &  &  &  & $\surd$ &  &  & $\surd$ & TC & PubMed20krct & 27.59 & 92.10 &   42.46 $\downarrow$\\
$\surd$ &  &  &  & $\surd$ &  &  & $\surd$ &  &  &  & $\surd$ & TC & PubMed20krct & 42.18 & 72.00 &   53.20 $\downarrow$\\
$\surd$ &  &  & $\surd$ &  &  &  &  & $\surd$ &  &  & $\surd$ & TC & PubMed20krct & 55.82 & 63.80 &   59.54 $\downarrow$\\
 & $\surd$ &  &  & $\surd$ &  & $\surd$ &  &  &  &  & $\surd$ & TC & PubMed20krct & 81.71 & 82.20 &   81.95 $\downarrow$\\
 & $\surd$ &  & $\surd$ &  &  & $\surd$ &  &  &  &  & $\surd$ & TC & PubMed20krct & 80.84 & 81.00 &   80.92 $\downarrow$\\
\bottomrule
\end{tabular}}
\label{table:TC_result}
\end{table*}
\end{center}